\documentclass[10pt,twocolumn,letterpaper]{article}

\usepackage{graphicx}
\usepackage{subcaption}
\usepackage{float}
\usepackage[justification=raggedright]{caption}	
\usepackage{lscape}                                         
\usepackage{wrapfig}

\usepackage[lined,ruled,linesnumbered]{algorithm2e}
\usepackage{animate} 

\usepackage{booktabs}                   
\usepackage{multirow}

\usepackage{makecell}

\usepackage{paralist}
\usepackage{enumitem}

\usepackage{bm}                          
\usepackage{epsfig}                      
\usepackage{graphicx}                  
\usepackage{times}
\usepackage{mathptmx}
\usepackage{mathtools}
\usepackage{amssymb,amsmath}   

\usepackage{units}
\usepackage{color}

\usepackage{comment}

\usepackage{url}  
\usepackage[pagebackref=true,breaklinks=true,letterpaper=true,colorlinks,bookmarks=false]{hyperref}
\usepackage{xspace}
\usepackage[table]{xcolor}
\usepackage{setspace}
\usepackage{grfext}
\PrependGraphicsExtensions*{.jpg,.png,.PNG}

\def\naive{na{\"i}ve\xspace}
\def\Naive{Na{\"i}ve\xspace}

\newlength\paramargin
\newlength\figmargin

\newlength\secmargin
\newlength\figcapmargin
\newlength\tabcapmargin

\newcommand{\red}{\textcolor{red}}
\newcommand{\blue}{\textcolor{blue}}

\newcommand{\heading}[1]
{
\vspace{1mm}\noindent\textbf{#1}
}

\newcommand{\secref}[1]{Section~\ref{sec:#1}}
\newcommand{\figref}[1]{Figure~\ref{fig:#1}} 
\newcommand{\tabref}[1]{Table~\ref{tab:#1}}

\long\def\ignorethis#1{}
\newcommand {\jiabin}[1]{{\color{cyan}\textbf{Jia-Bin: }#1}\normalfont}

\newcommand{\tb}[1]{\textbf{#1}}

\newbox\jsavebox%

\def\xi{\mathbf{x}_i}

\graphicspath{{figure}, {example}}

\usepackage{cvpr}
\usepackage{animate}
\usepackage{subfiles} 
\cvprfinalcopy 

\def\cvprPaperID{6222} 

\ifcvprfinal\fi

\begin{document}

\title{3D Photography using Context-aware Layered Depth Inpainting}

\author{Meng-Li Shih$^{12}$\\{\tt\small shihsml@gapp.nthu.edu.tw}
        \and Shih-Yang Su$^{1}$\\{\tt\small shihyang@vt.edu}
        \and Johannes Kopf$^3$\\{\tt\small jkopf@fb.com}
        \and Jia-Bin Huang$^1$\\{\tt\small jbhuang@vt.edu} 
        \and $^{1}$Virginia Tech 
        \and $^{2}$National Tsing Hua University 
        \and $^{3}$Facebook
        \and \tt\small \url{https://shihmengli.github.io/3D-Photo-Inpainting}}
        


\twocolumn[{
\renewcommand\twocolumn[1][]{#1}
\maketitle
\vspace{-1.0cm}
\vspace{-3mm}
\newlength\fta
\setlength\fta{4.2cm}
\begin{center}
   \centering%
\parbox[t]{\fta}{\vspace{0mm}\centering%
  \includegraphics[width=\fta,trim=40 120 40 100,clip]{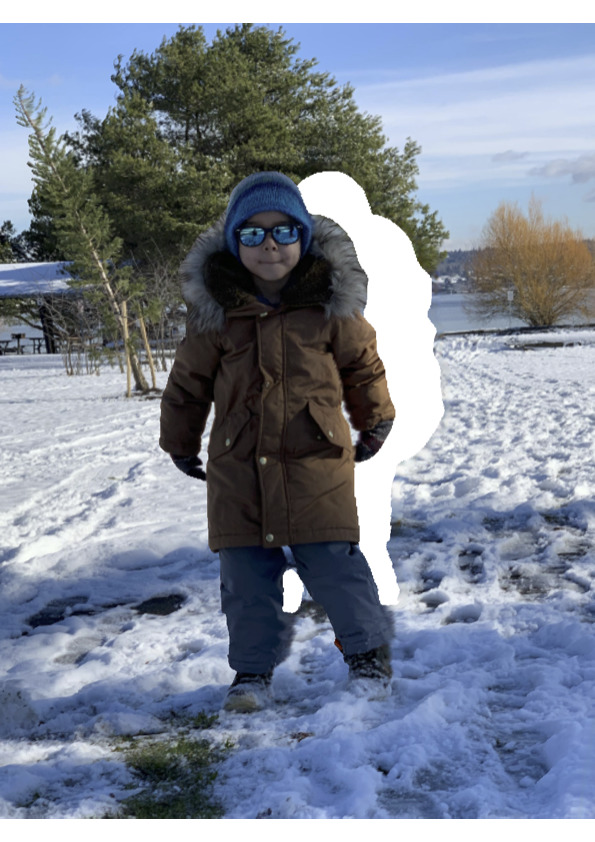}\vspace{-0.5mm}\\%
	{\small (a) Depth-warping  (holes)}}%
\hfill%
\parbox[t]{\fta}{\vspace{0mm}\centering%
  \includegraphics[width=\fta,trim=40 120 40 100,clip]{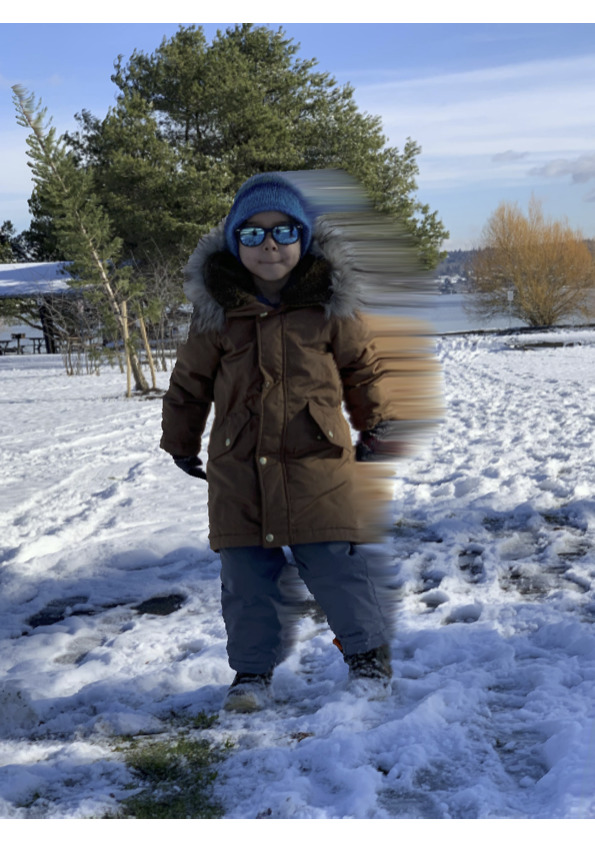}\vspace{-0.5mm}\\%
	{\small(b) Depth-warping (stretching)}}%
\hfill%
\parbox[t]{\fta}{\vspace{0mm}\centering%
  \includegraphics[width=\fta,trim=40 120 40 100,clip]{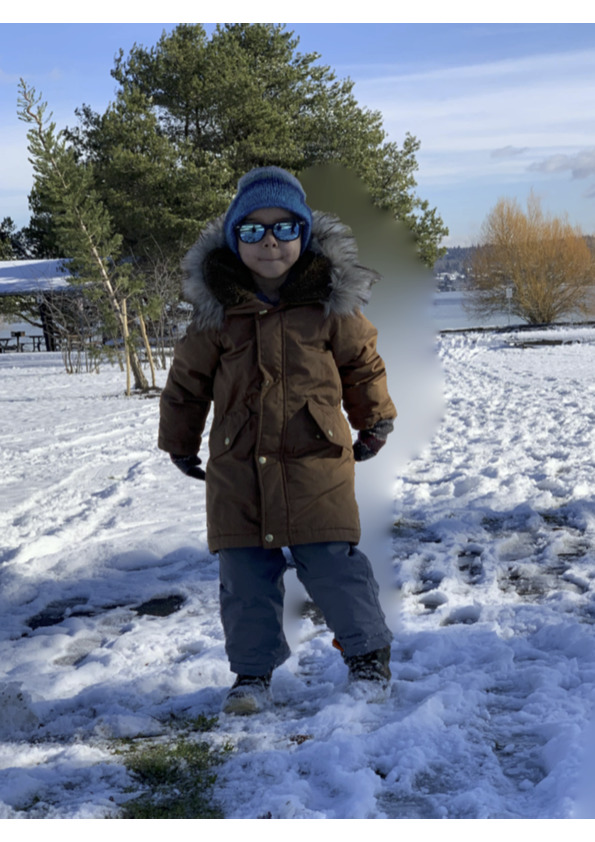}\vspace{-0.5mm}\\%
	{\small(c) Facebook 3D photo}}%
\hfill%
\parbox[t]{\fta}{\vspace{0mm}\centering%
  \includegraphics[width=\fta,trim=40 120 40 100,clip]{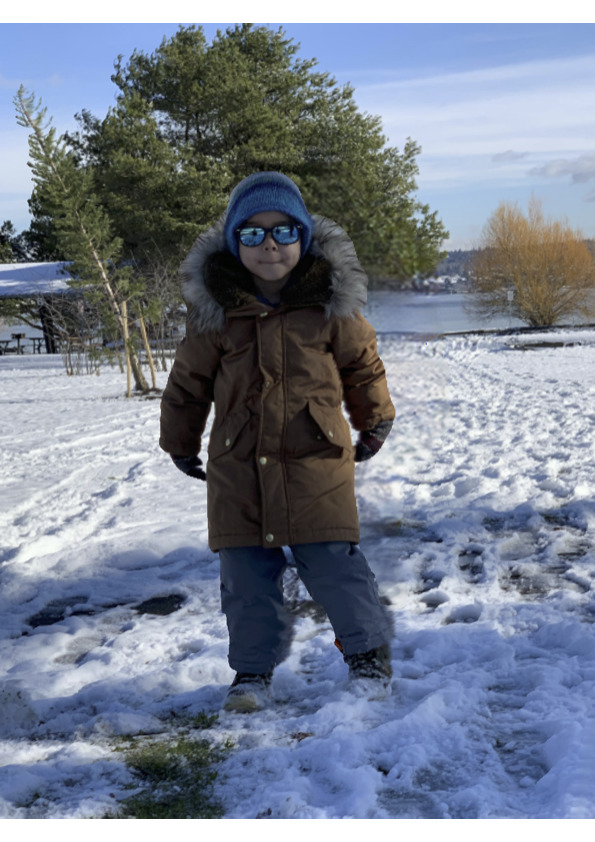}\vspace{-0.5mm}\\%
	{\small (d) Our result}}%
\vspace{1.5mm}\\
\captionof{figure}{
\tb{3D photography from a single RGB-D image.}
\Naive methods either produce holes (a) or stretch content (b) at disocclusions.
Color and depth inpainting using diffusion is better, but provides a too smooth appearance (c).
Our approach is capable of synthesizing new color/depth texture and structures, leading to more photorealistic novel views (d).
}
\label{fig:teaser}
\end{center}

}]

\thispagestyle{empty}
\begin{abstract}
We propose a method for converting a single RGB-D input image into a \emph{3D photo} --- a multi-layer representation for novel view synthesis that contains hallucinated color and depth structures in regions occluded in the original view.
We use a Layered Depth Image with explicit pixel connectivity as underlying representation, and present a learning-based inpainting model that synthesizes new local color-and-depth content into the occluded region in a spatial context-aware manner.
The resulting 3D photos can be efficiently rendered with motion parallax using standard graphics engines.
We validate the effectiveness of our method on a wide range of challenging everyday scenes and show less artifacts compared with the state of the arts.
\vspace{-3mm}
\end{abstract}
\section{Introduction}
\label{sec:intro}
3D photography---capturing views of the world with a camera and using image-based rendering techniques for novel view synthesis---is a fascinating way to record and reproduce visual perception.
It provides a dramatically more immersive experience than old 2D photography: almost lifelike in Virtual Reality, and even to some degree on normal flat displays when displayed with parallax.

Classic image-based reconstruction and rendering techniques, however, require elaborate capture setups involving many images with large baselines \cite{hedman2017casual,whelan2018reconstructing,kopf2013image,penner2017soft,hedman2018deep,flynn2016deepstereo}, and/or special hardware (e.g., Lytro Immerge, Facebook Manifold camera\footnote{\url{https://facebook360.fb.com/2018/05/01/red-facebook-6dof-camera/}}).

Recently, we have seen work to make capture for 3D photography more effortless by using cell phone cameras and lowering baseline requirements \cite{hedman2017casual,hedman2018instant}.
In the most extreme cases, novel techniques such as Facebook 3D Photos\footnote{\url{https://facebook360.fb.com/2018/10/11/3d-photos-now-rolling-out-on-facebook-and-in-vr/}} now just require capturing a \emph{single snapshot} with a dual lens camera phone, which essentially provides an RGB-D (color and depth) input image.

In this work we are interested in rendering novel views from such an RGB-D input.
The most salient features in rendered novel views are the disocclusions due to parallax:
\naive depth-based warping techniques either produce gaps here (Figure~\ref{fig:teaser}a)
or stretched content (\ref{fig:teaser}b).
Recent methods try to provide better extrapolations.

Stereo magnification~\cite{zhou2018stereo} and recent variants~\cite{srinivasan2019pushing,mildenhall2019llff} use a fronto-parallel multi-plane representation (MPI), which is synthesized from the small-baseline dual camera stereo input.
However, MPI produces artifacts on sloped surfaces. 
Besides, the excessive redundancy in the multi-plane representation makes it memory and storage inefficient and costly to render.

Facebook 3D Photos use a layered depth image (LDI) representation \cite{shade1998layered}, which is more compact due to its sparsity, and can be converted into a light-weight mesh representation for rendering.
The color and depth in occluded regions are synthesized using heuristics that are optimized for fast runtime on mobile devices.
In particular it uses a isotropic diffusion algorithm for inpainting colors, which produces overly smooth results and is unable to extrapolate texture and structures (Figure~\ref{fig:teaser}c).

Several recent learning-based methods also use similar multi-layer image representations \cite{dhamo2018peeking,tulsiani2018layer}.
However, these methods use ``rigid'' layer structures, in the sense that every pixel in the image has the same (fixed and predetermined) number of layers.
At every pixel, they store the nearest surface in the first layer, the second-nearest in the next layer, etc.
This is problematic, because across depth discontinuities the content within a layer changes abruptly, which destroys locality in receptive fields of convolution kernels.

In this work we present a new learning-based method that generates a 3D photo from an RGB-D input.
The depth can either come from dual camera cell phone stereo, or be estimated from a single RGB image~\cite{li2018megadepth,lasinger2019towards,godard2019digging}.
We use the LDI representation (similar to Facebook 3D Photos) because it is compact and allows us to handle situations of arbitrary depth-complexity.
Unlike the ``rigid" layer structures described above, we explicitly store connectivity across pixels in our representation.
However, as a result it is more difficult to apply a \emph{global} CNN to the problem, because our topology is more complex than a standard tensor.
Instead, we break the problem into many local inpainting sub-problems, which we solve iteratively.
Each problem is locally like an image, so we can apply standard CNN.
We use an inpainting model that is conditioned on \emph{spatially-adaptive} context regions, which are extracted from the local connectivity of the LDI.
After synthesis we fuse the inpainted regions back into the LDI, leading to a recursive algorithm that proceeds until all depth edges are treated.

The result of our algorithm are 3D photos with synthesized texture and structures in occluded regions (Figure~\ref{fig:teaser}d).
Unlike most previous approaches we do not require predetermining a fixed number of layers.
Instead our algorithm adapts by design to the local depth-complexity of the input and generates a varying number of layers across the image.
We have validated our approach on a wide variety of photos captured in different situations.

\ignorethis{
Image-based rendering (IBR) techniques provide free-viewpoint rendering of the captured scene, providing the users with immersive experiences with 3D motion parallax. Existing IBR techniques, however, either require special hardwares (e.g., Lytro Immerge), depth sensors~\cite{hedman2016scalable}, or many images with sufficiently large baselines for a reliable 3D scene reconstruction~\cite{hedman2017casual,whelan2018reconstructing,kopf2013image,penner2017soft,hedman2018deep,flynn2016deepstereo}.

Much of the recent efforts have been devoted to make novel view synthesis applicable to wider scenarios using narrow baseline stereo inputs (e.g., captured with cellphones with a dual camera)~\cite{zhou2018stereo,choi2018extreme} or using one single image~\cite{dhamo2018peeking,tulsiani2018layer,facebook3D}. While impressive results have been shown, existing approaches have difficulty in handling sloped surfaces~\cite{zhou2018stereo}, inferring complex multi-layer structures~\cite{dhamo2018peeking,tulsiani2018layer}, or producing plausible contents for the occluded regions~\cite{hedman2018instant,facebook3D}.

In this work, we present an algorithm that can take a everyday photograph as input and produce a highly detailed layered depth image that support free-viewpoint rendering. We start with estimating the depth of the given input color image with single-view depth estimation model~\cite{li2018megadepth} or directly using the reconstructed depth from dual lens cell phone cameras. We then localize and link edges in the depth map. For each linked depth discontinuity, we apply an learning-based inpaint model to fill up the color, depth, and depth discontinuities of the occluded regions (i.e., in one side of the depth discontinuity) conditioned on spatially adaptive contexts. Such a localized, context-aware inpainting allows us to infer complex multi-layer structure and contents of the invisible layers. We fuse all the inpainted layers with the visible layers into a connected, layered depth image. We validate our approach on a wide variety of photos captured in everyday scenes. Our results show that our algorithm can create compelling novel view rendering with considerably fewer visual artifacts than existing techniques. 

\paragraph{Our contributions.} 
\begin{itemize}
\item We present a fully automatic algorithm for enabling high-quality novel view synthesis \emph{from a single image}. Unlike existing learning-based view synthesis techniques that focus on specific domains (e.g., street view~\cite{tulsiani2018layer,liu2018geometry} or synthetic scenes~\cite{dhamo2018peeking}, or objects~\cite{zhou2016view,park2017transformation}), we show that our method is applicable to a wide variety of challenging scenes.
\item We propose to recover both the structure and contents of the occluded regions through localized inpainting with spatially adaptive contexts. We design our inpainting network so that both the recovered depth and color respect the inferred depth discontinuity. This allows us to move beyond handling simple scenes with clear foreground/background layer separation. 
\end{itemize}
}











\section{Related Work}
\label{sec:related}


\heading{Representation for novel view synthesis.}
Different types of representations have been explored for novel view synthesis, including light fields~\cite{gortler1996lumigraph,levoy1996light,buehler2001unstructured}, multi-plane images~\cite{zhou2018stereo,srinivasan2019pushing,mildenhall2019llff}, and layered depth images~\cite{shade1998layered,swirski2011layered,dhamo2018peeking,tulsiani2018layer,hedman2017casual,hedman2018instant,dhamo2019object,niklaus2019KenBurns}. 
Light fields enable photorealistic rendering of novel views, but generally require many input images to achieve good results.
The multi-plane image representation~\cite{zhou2018stereo,srinivasan2019pushing,mildenhall2019llff} stores multiple layers of RGB-$\alpha$ images at fixed depths. The main advantage of this representation is its ability to capture semi-reflective or semi-transparent surfaces. 
However, due to the fixed depth discretization, sloped surfaces often do not reproduce well, unless an excessive number of planes is used.
Many variants of layered depth image representations have been used over time.
Representations with a fixed number of layers everywhere have recently been used \cite{dhamo2018peeking,tulsiani2018layer}, but they do not preserve locality well, as described in the previous section.
Other recent work~\cite{hedman2017casual,hedman2018instant} extends the original work of Shade et al.~\cite{shade1998layered} to explicitly store connectivity information.
This representation can locally adapt to any depth-complexity and can be easily converted into a textured mesh for efficient rendering.
Our work uses this representation as well.



\heading{Image-based rendering.} 
Image-based rendering techniques enable photorealistic synthesis of novel views from a collection of posed images. 
These methods work best when the images have sufficiently large baselines (so that multi-view stereo algorithms can work well) or are captured with depth sensors. Recent advances include learning-based blending~\cite{hedman2018deep}, soft 3D reconstruction~\cite{penner2017soft}, handling reflection~\cite{sinha2012image,kopf2013image}, relighting~\cite{xu2019deep}, and reconstructing mirror and glass surfaces~\cite{whelan2018reconstructing}. 
Our focus in this work lies in novel view synthesis \emph{from one single image}.


\heading{Learning-based view synthesis.}
CNN-based methods have been applied to synthesizing novel views from sparse light field data~\cite{kalantari2016learning} or two or more posed images~\cite{flynn2016deepstereo,hedman2018deep,choi2019extreme}. 
Several recent methods explore view synthesis from a single image. 
These methods, however, often focus on a specific domain~\cite{srinivasan2017learning,Wiles20SynSin}, synthetic 3D scenes/objects~\cite{zhou2016view,park2017transformation,sun2018multi,dhamo2019object,dhamo2018peeking,eslami2018neural}, hallucinating only one specific view~\cite{xie2016deep3d,zeng2015hallucinating}, or assuming piecewise planar scenes~\cite{liu2018planenet,liu2018geometry}.

Many of these learning-based view synthesis methods require running a forward pass of the pre-trained network to synthesize the image of a given viewpoint. 
This makes these approaches less applicable to display on resource-constrained devices.
Our representation, on the other hand, can be easily converted into a textured mesh and efficiently rendered with standard graphics engines.





\heading{Image inpainting.} 
The task of image inpainting aims to fill missing regions in images with plausible content. 
Inspired by the success of texture synthesis~\cite{efros1999texture,efros2001image}, \emph{example-based methods} complete the missing regions by transferring the contents from the known regions of the image, either through non-parametric patch-based synthesis~\cite{wexler2007space,barnes2009patchmatch,darabi2012image,huang2014image} or solving a Markov Random Field model using belief propagation~\cite{komodakis2007image} or graph cut~\cite{pritch2009shift,kwatra2003graphcut,he2014image}.
Driven by the progress of convolutional neural networks, \emph{CNN-based methods} have received considerable attention due to their ability to predict semantically meaningful contents that are not available in the known regions~\cite{pathak2016context,song2017contextual,iizuka2017globally,yang2017high,yu2018generative}. Recent efforts include designing CNN architectures to better handle holes with irregular shapes~\cite{liu2018image,yu2019free,Yan_2018_Shift} and two-stage methods with structure-content disentanglement, e.g., predicting structure (e.g., contour/edges in the missing regions) and followed by content completion conditioned on the predicted structures~\cite{nazeri2019edgeconnect,xiong2019foreground,ren2019structureflow}.

Our inpainting model builds upon the recent two-stage approaches~\cite{nazeri2019edgeconnect,xiong2019foreground,ren2019structureflow} but with two key differences. First, unlike existing image inpainting algorithms where the hole and the available contexts are \emph{static} (e.g., the known regions in the entire input image), we apply the inpainting \emph{locally} around each depth discontinuity with \emph{adaptive} hole and context regions. Second, in addition to inpaint the color image, we also inpaint the depth values as well as the depth discontinuity in the missing regions.

\heading{Depth inpainting.} 
Depth inpainting has applications in filling missing depth values where commodity-grade depth cameras fail (e.g., transparent/reflective/distant surfaces)~\cite{liu2017robust,zhang2018deep,lu2014depth} or performing image editing tasks such as object removal on stereo images~\cite{wang2008stereoscopic,mu2014stereoscopic}. 
The goal of these algorithms, however, is to inpaint the depth of the \emph{visible surfaces}. In contrast, our focus is on recovering the depth of the \emph{hidden surface}.

\heading{CNN-based single depth estimation.} 
CNN-based methods have recently demonstrated promising results on estimating depth from a single image. Due to the difficulty of collecting labeled datasets, earlier approaches often focus on specific visual domains such as indoor scenes~\cite{eigen2015predicting} or street view~\cite{godard2017unsupervised,zhou2017unsupervised}. 
While the accuracy of these approaches is not yet competitive with multi-view stereo algorithms, this line of research is particularly promising due to the availability of larger and more diverse training datasets from relative depth annotations~\cite{chen2016single}, multi-view stereo~\cite{li2018megadepth}, 3D movies~\cite{lasinger2019towards} and synthetic data~\cite{niklaus2019KenBurns}.

For cases where only one single color image is available, we obtain the depth estimate through a pre-trained depth estimation model~\cite{li2018megadepth,lasinger2019towards}. 
Removing the dependency on stereo or multiple images as input makes our method more widely applicable to all the existing photos.

\section{Method}
\label{sec:method}

\heading{Layered depth image.}
Our method takes as input an RGB-D image (i.e., an aligned color-and-depth image pair) and generates a \emph{Layered Depth Image} (LDI, \cite{shade1998layered}) with inpainted color and depth in parts that were occluded in the input.

An LDI is similar to a regular 4-connected image, except at every position in the pixel lattice it can hold any number of pixels, from zero to many.
Each LDI pixel stores a color and a depth value.
Unlike the original LDI work \cite{shade1998layered}, we explicitly represent the local connectivity of pixels: each pixel stores pointers to either zero or at most one direct neighbor in each of the four cardinal directions (left, right, top, bottom).
LDI pixels are 4-connected like normal image pixels within smooth regions, but do not have neighbors across depth discontinuities.

LDIs are a useful representation for 3D photography, because
(1) they naturally handle an arbitrary number of layers, i.e., can adapt to depth-complex situations as necessary, and 
(2) they are sparse, i.e., memory and storage efficient and can be converted into a light-weight textured mesh representation that renders fast.

The quality of the depth input to our method does not need to be perfect, as long as discontinuities are reasonably well aligned in the color and depth channels.
In practice, we have successfully used our method with inputs from dual camera cell phones as well as with estimated depth maps from learning-based methods \cite{li2018megadepth,lasinger2019towards}.

\heading{Method overview.}
Given an input RGB-D image, our method proceeds as follows.
We first initialize a trivial LDI, which uses a single layer everywhere and is fully 4-connected.
In a pre-process we detect major depth discontinuities and group them into simple connected \emph{depth edges} (\secref{preprocessing}).
These form the basic units for our main algorithm below.
In the core part of our algorithm, we iteratively select a depth edge for inpainting.
We then \emph{disconnect} the LDI pixels across the edge and only consider the background pixels of the edge for inpainting.
We extract a local \emph{context region} from the ``known'' side of the edge, and generate a \emph{synthesis region} on the ``unknown'' side (\secref{context_syn_region}).
The synthesis region is a contiguous 2D region of \emph{new} pixels, whose color and depth values we generate from the given context using a learning-based method (\secref{inpainting}).
Once inpainted, we merge the synthesized pixels back into the LDI (\secref{convert_to_3d}).
Our method iteratively proceeds in this manner until all depth edges have been treated.

\newlength\flea
\setlength\flea{2.7cm}
\newlength\fleb
\setlength\fleb{3.75cm}
\newlength\flec
\setlength\flec{1.60cm}

\begin{figure*}[t]
\parbox[t]{\fleb}{\vspace{0mm}\centering%
  \includegraphics[width=\fleb,trim=0 150 0 150,clip]%
	    {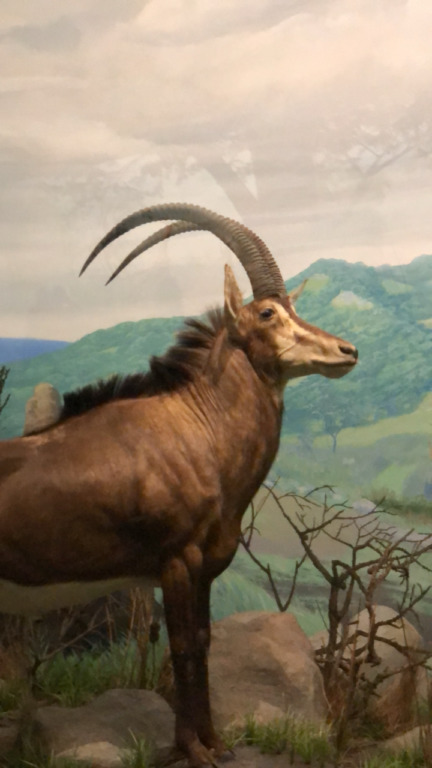}\vspace{-1mm}\\%
	(a) Color%
}%
\hfill%
\parbox[t]{\fleb}{\vspace{0mm}\centering%
  \includegraphics[width=\fleb,trim=0 150 0 150,clip]%
	    {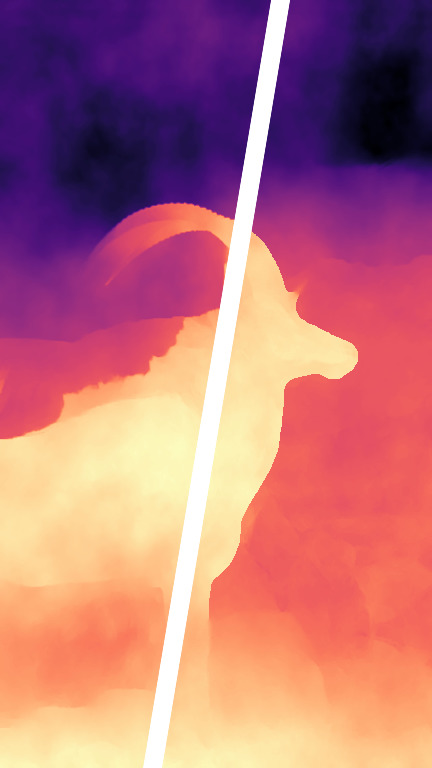}\vspace{-1mm}\\%
	(b) Raw / filtered depth%
}%
\hfill%
\parbox[t]{\flec}{\vspace{0mm}\centering%
  \fbox{\includegraphics[width=\flec]%
	    {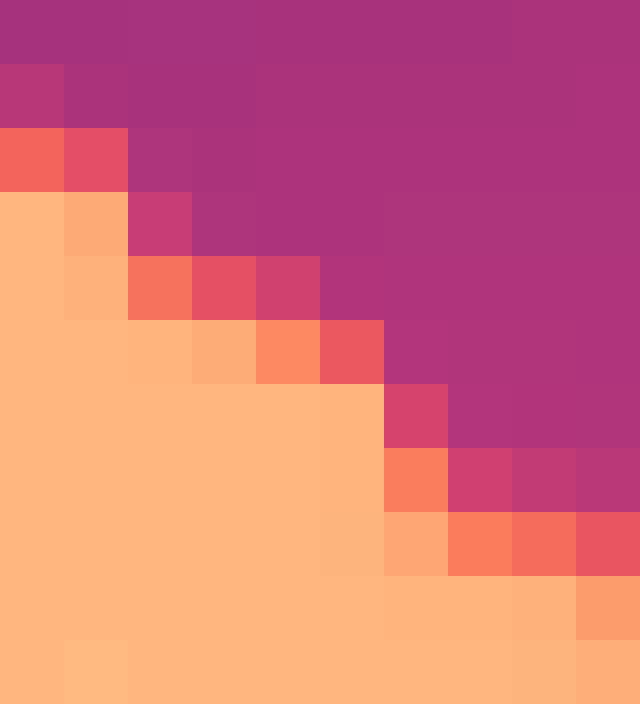}}\vspace{-1mm}\\%
	(c) Raw\vspace{1mm}\\%
  \fbox{\includegraphics[width=\flec]%
	    {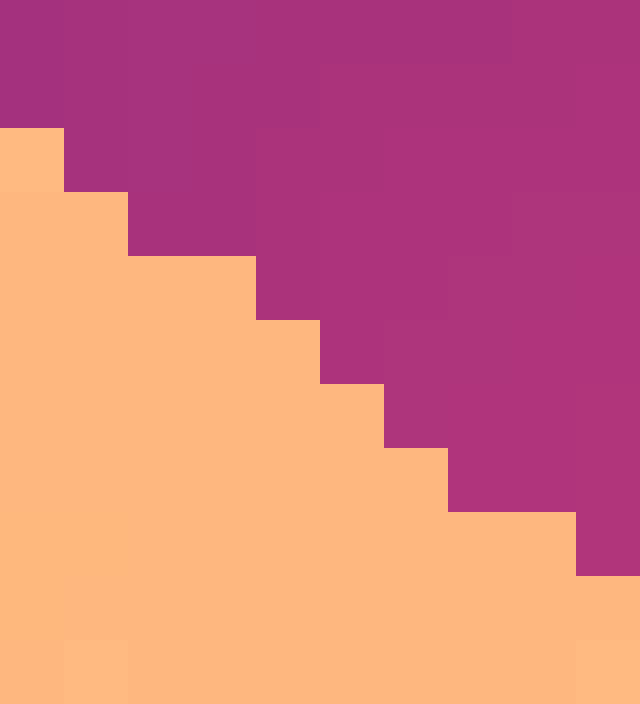}}\vspace{-1mm}\\%
	(d) Filtered%
}%
\hfill%
\parbox[t]{\fleb}{\vspace{0mm}\centering%
  \fbox{\includegraphics[width=\fleb,trim=0 300 0 300,clip]%
	    {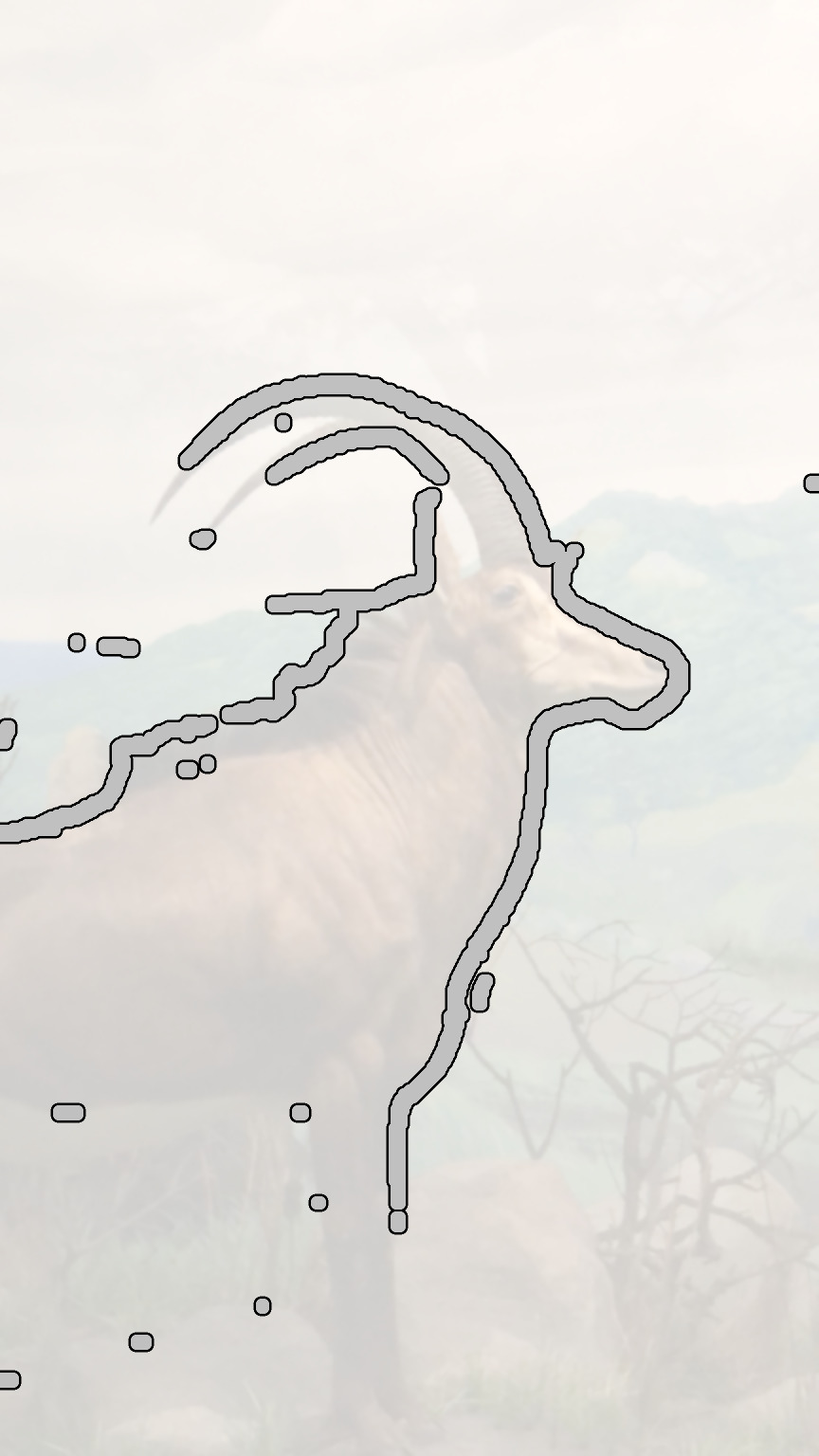}}\vspace{-1mm}\\%
	(e) Raw discontinuities%
}%
\hfill%
\parbox[t]{\fleb}{\vspace{0mm}\centering%
  \fbox{\includegraphics[width=\fleb,trim=0 300 0 300,clip]%
	    {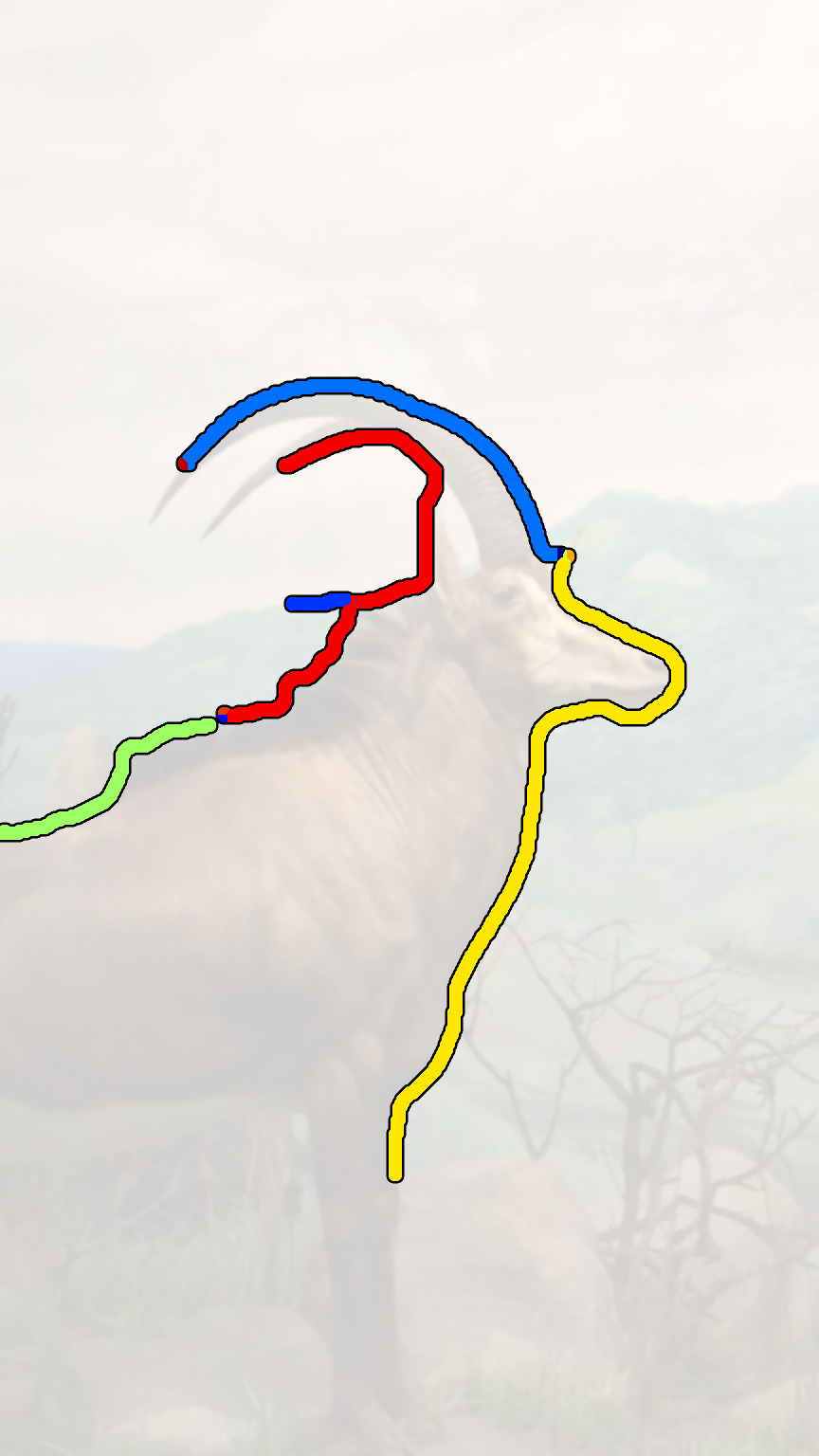}}\vspace{-1mm}\\%
	(f) Linked depth edges%
}%
\vspace{\figcapmargin}
\caption{
\tb{Preprocessing.} Preprocessing of the color and depth input (a-b).
We use a bilateral median filter to sharpen the input depth maps (c-d),
detect raw discontinuities using disparity thresholds (e), and clean up spurious threshold responses and link discontinuities into connected depth edges (f).
These linked depth edges form the basic unit for our inpainting process.
}
\label{fig:edges}
\end{figure*}
\newlength\flla
\setlength\flla{4.1cm}

\begin{figure*}[t]
\parbox[t]{\flla}{\vspace{0mm}\centering%
  \includegraphics[width=\flla]{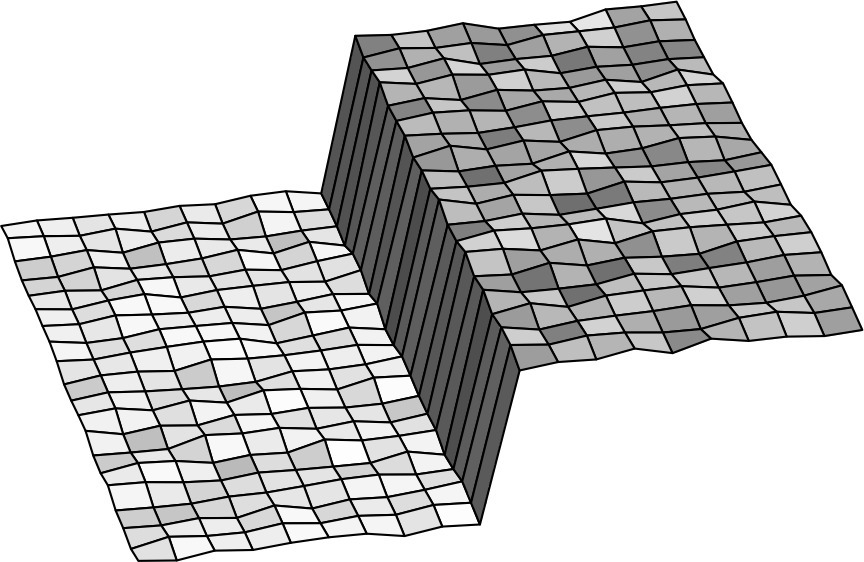}\vspace{-1mm}\\%
	(a) Initial LDI\\
	~~~~~(fully connected)%
}%
\hfill%
\parbox[t]{\flla}{\vspace{0mm}\centering%
  \includegraphics[width=\flla]{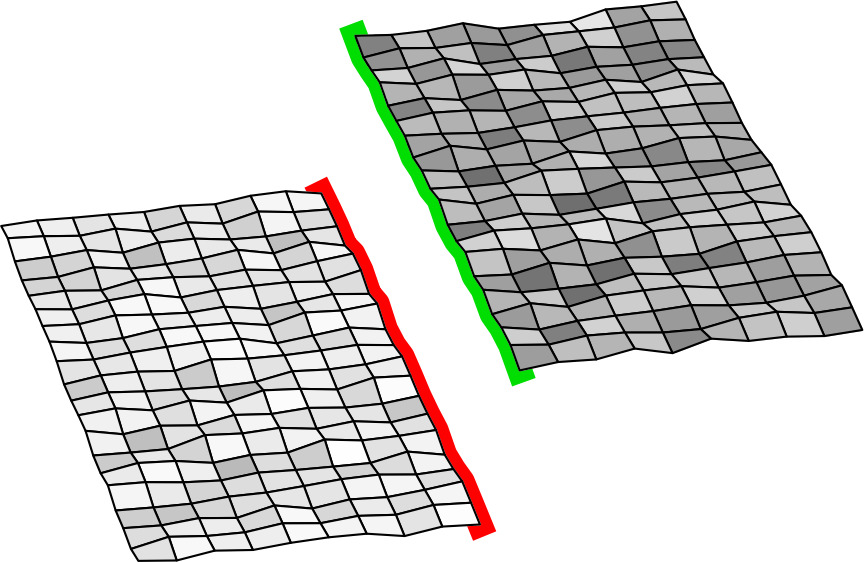}\vspace{-1mm}\\%
	(b) Cut across discontinuity%
}%
\hfill%
\parbox[t]{\flla}{\vspace{0mm}\centering%
  \includegraphics[width=\flla]{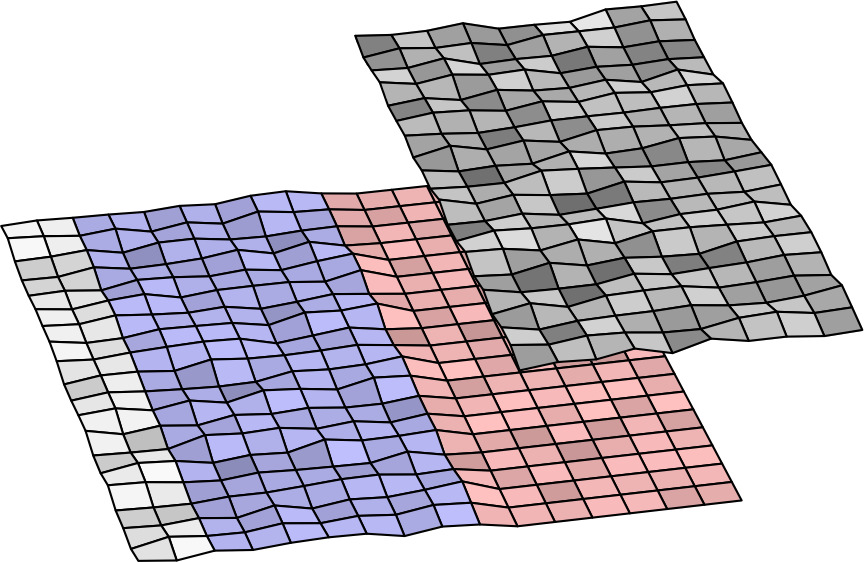}\vspace{-1mm}\\%
	(c) Context / synthesis regions%
}%
\hfill%
\parbox[t]{\flla}{\vspace{0mm}\centering%
  \includegraphics[width=\flla]{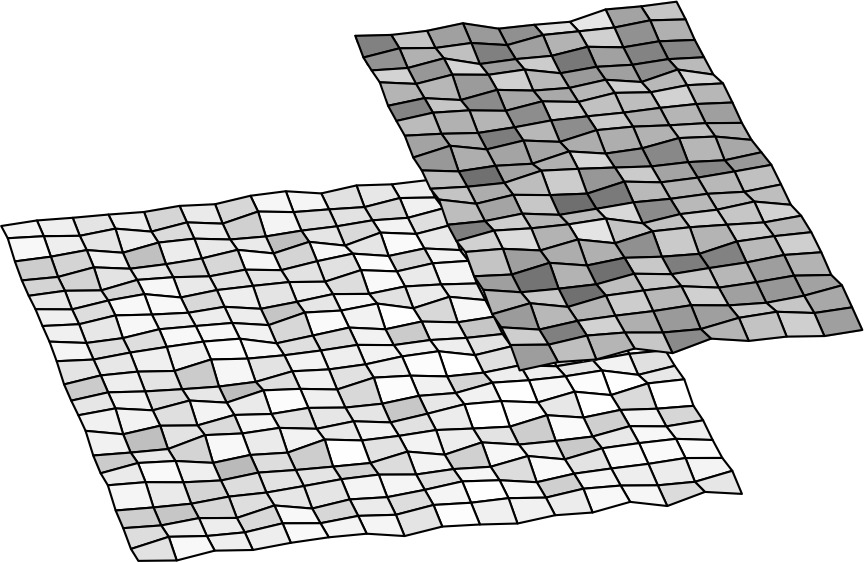}\vspace{-1mm}\\%
	(d) Inpainted%
}%
\vspace{\figcapmargin}
\caption{
\tb{Conceptual illustration of the LDI inpainting algorithm.}
(a) The initial LDI is fully connected. A depth edge (discontinuity) is marked in gray.
(b) We first cut the LDI pixel connections across the depth, forming a foreground silhouette ({\color{green}green}) and a background silhouette (\red{red}).
(c) For the background silhouette we spawn a context region (\blue{blue}) and a synthesis region (\red{red}) of new LDI pixels.
(d) The synthesized pixels have been merged into the LDI.
}
\vspace{-3mm}
\label{fig:ldi}
\end{figure*}

\subsection{Image preprocessing}
\label{sec:preprocessing}

The only input to our method is a single RGB-D image.
Every step of the algorithm below proceeds fully automatically.
We normalize the depth channel, by mapping the min and max disparity values (i.e., 1 / depth) to 0 and 1, respectively.
All parameters related to spatial dimensions below are tuned for images with 1024 pixels along the longer dimension, and should be adjusted proportionally for images of different sizes.  

We start by lifting the image onto an LDI, i.e., creating a single layer everywhere and connecting every LDI pixel to its four cardinal neighbors.
Since our goal is to inpaint the occluded parts of the scene, we need to find \emph{depth discontinuities} since these are the places where we need to extend the existing content.
In most depth maps produced by stereo methods (dual camera cell phones) or depth estimation networks, discontinuities are blurred across multiple pixels (Figure~\ref{fig:edges}c), making it difficult to precisely localize them.
We, therefore, sharpen the depth maps using a bilateral median filter \cite{ma2013median} (Figure~\ref{fig:edges}d), using a $7 \!\times\! 7$ window size, and $\sigma_\textit{spatial} = 4.0$, $\sigma_\textit{intensity} = 0.5$.

After sharpening the depth map, we find discontinuities by thresholding the disparity difference between neighboring pixels.
%
%
This results in many spurious responses, such as isolated speckles and short segments dangling off longer edges (Figure~\ref{fig:edges}e).
We clean this up as follows:
First, we create a binary map by labeling depth discontinuities as 1 (and others as 0).
Next, we use connected component analysis to merge adjacent discontinuities into a collection of ``linked depth edges".  
To avoid merging edges at junctions, we separate them based on the local connectivity of the LDI.  
Finally, we remove short segments ($<10$ pixels), including both isolated and dangling ones. 
We determine the threshold 10 by conducting five-fold cross-validation with LPIPS~\cite{zhang2018unreasonable} metric on 50 samples randomly selected from RealEstate10K training set.
The final edges (Figures~\ref{fig:edges}f) form the basic unit of our iterative inpainting procedure, which is described in the following sections.

\newlength\flca
\setlength\flca{2.7cm}

\begin{figure}[t]
\fbox{\includegraphics[width=\flca,trim=0 350 0 200,clip]{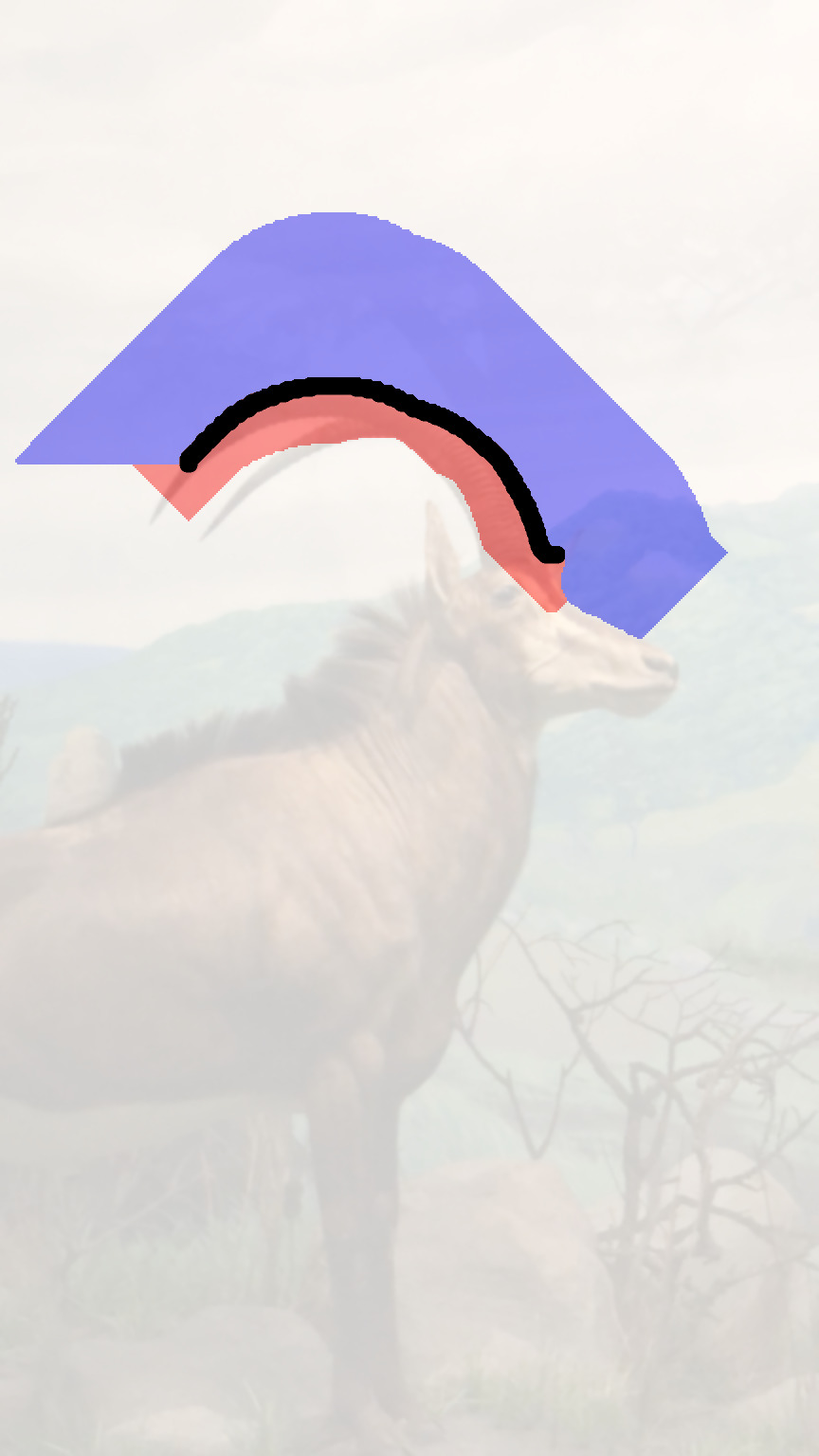}}%
\hfill%
\fbox{\includegraphics[width=\flca,trim=0 350 0 200,clip]{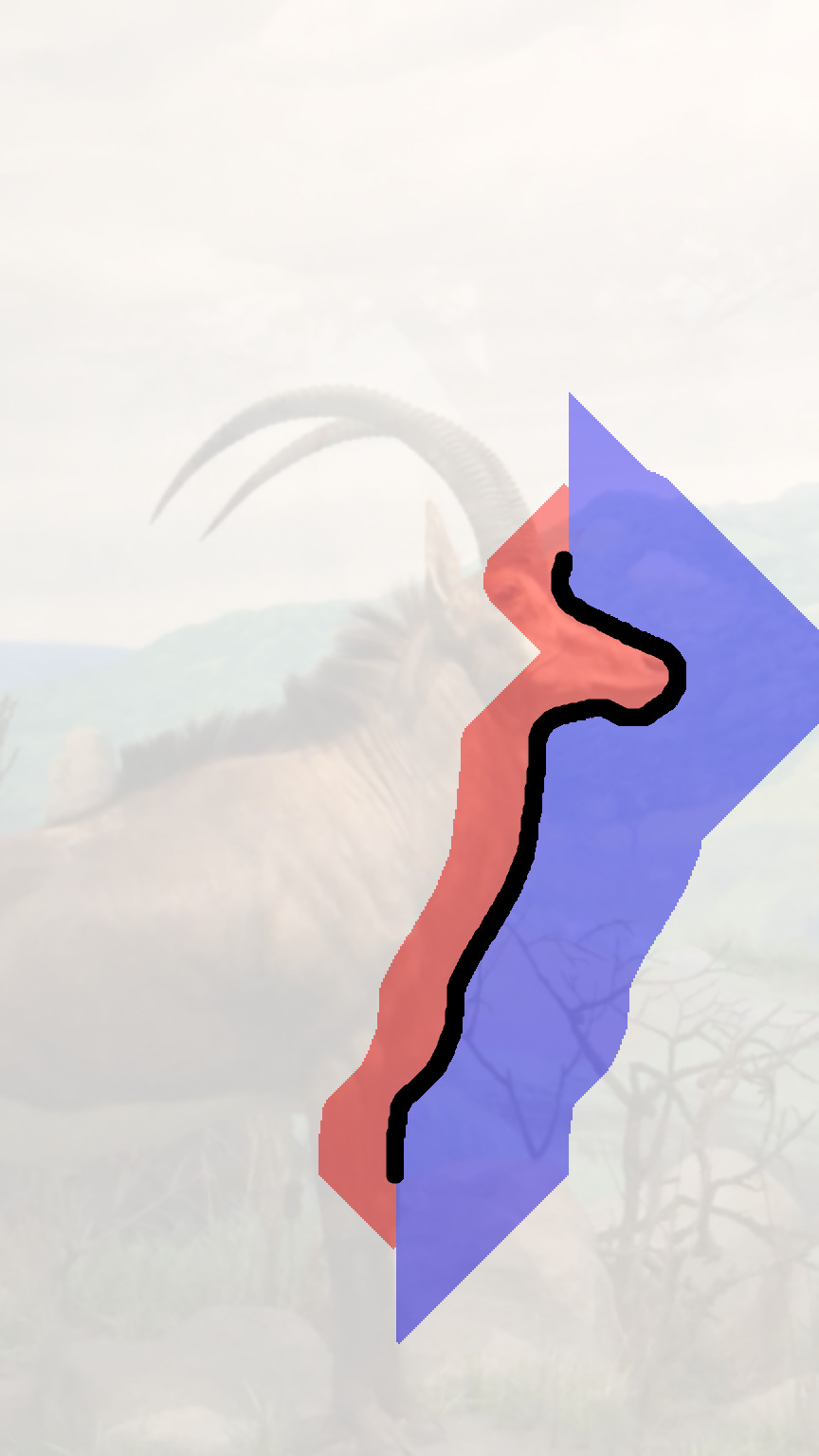}}%
\hfill%
\fbox{\includegraphics[width=\flca,trim=0 350 0 200,clip]{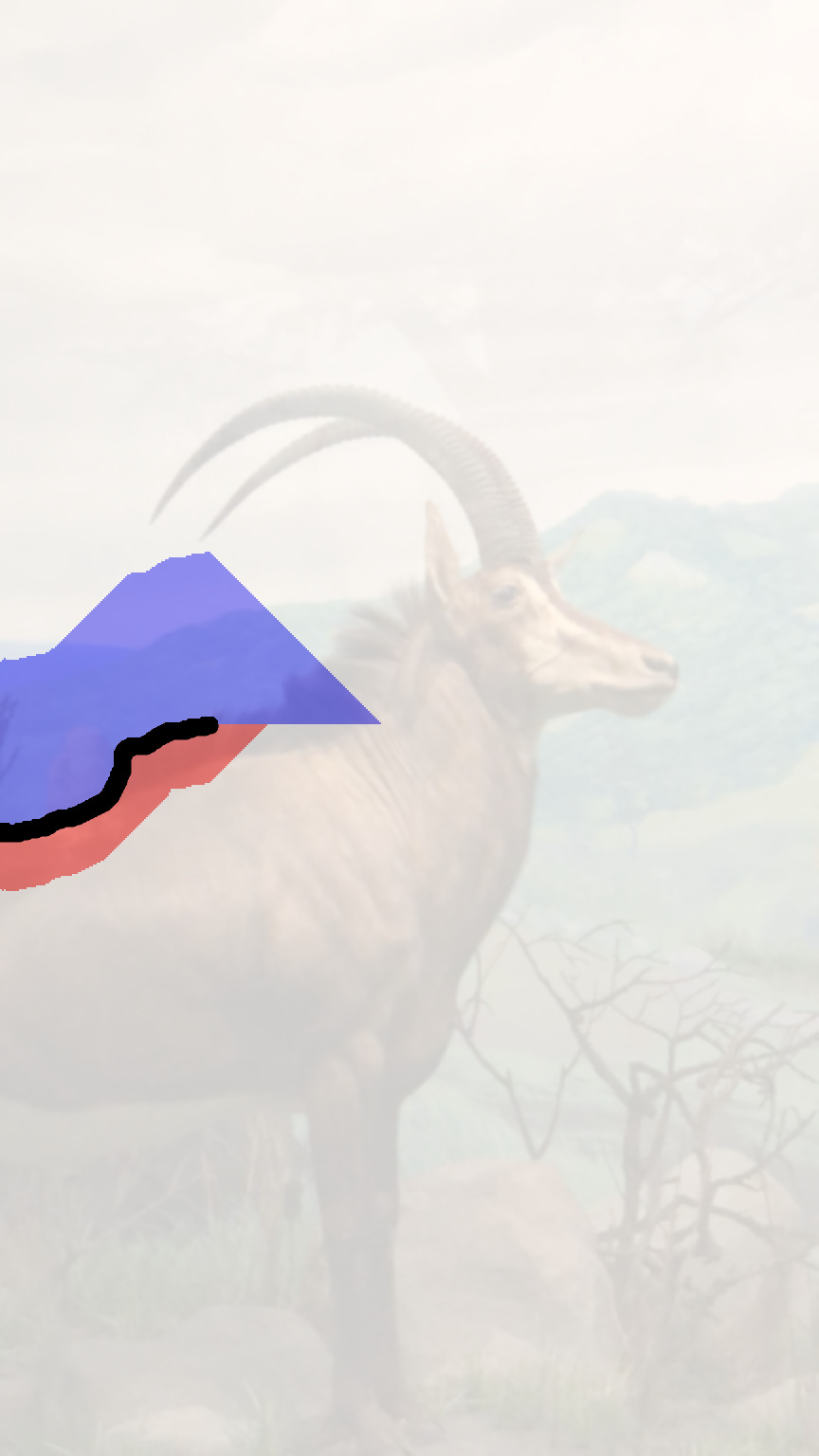}}%
\vspace{\figcapmargin}
\caption{\tb{Context/synthesis regions.} 
Context regions (\blue{blue}) and synthesis regions (\red{red}) for three example connected depth edges (black) from Figure~\ref{fig:edges}(f).
}
\label{fig:context_example}
\end{figure}
\newlength\ee
\setlength\ee{1.956cm}
\begin{figure}[t]
\parbox[t]{\ee}{\vspace{0mm}\centering%
  \includegraphics[trim=180 150 180 50,clip,width=\ee]{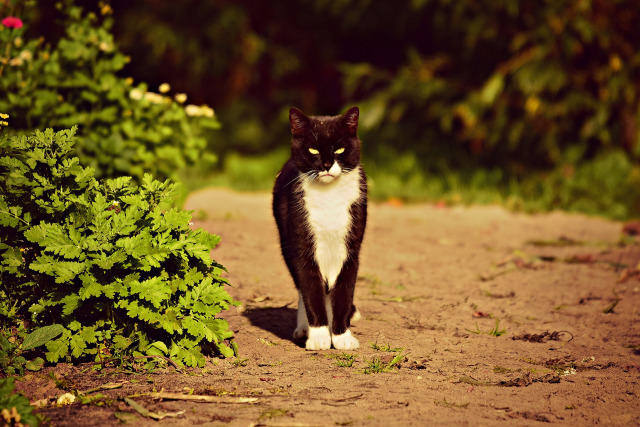}
  {\small\vspace{-4mm}\\Input}%
}\hfill%
\parbox[t]{\ee}{\vspace{0mm}\centering%
  \includegraphics[trim=180 150 180 50,clip,width=\ee]{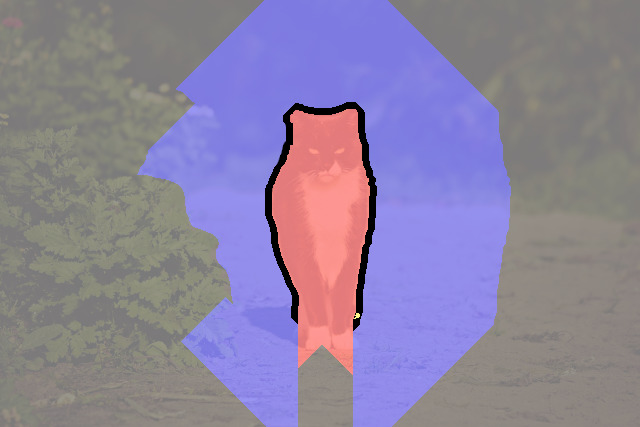}
  {\small \vspace{-4mm}\\context/synthesis}%
}\hfill%
\parbox[t]{\ee}{\vspace{0mm}\centering%
  \includegraphics[trim=180 150 180 50,clip,width=\ee]{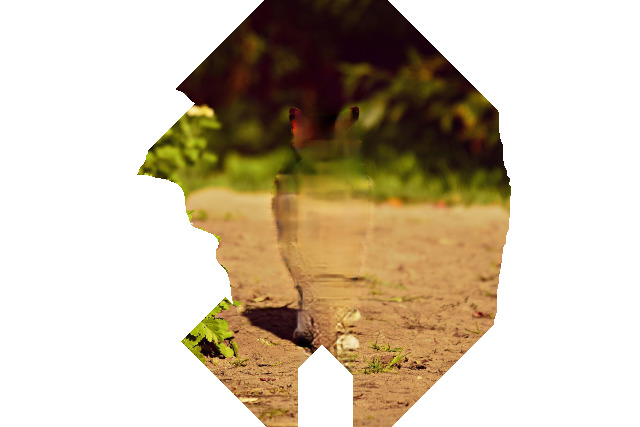}
  {\small \vspace{-4mm}\\w/o dilation}%
}\hfill%
\parbox[t]{\ee}{\vspace{0mm}\centering%
  \includegraphics[trim=180 150 180 50,clip,width=\ee]{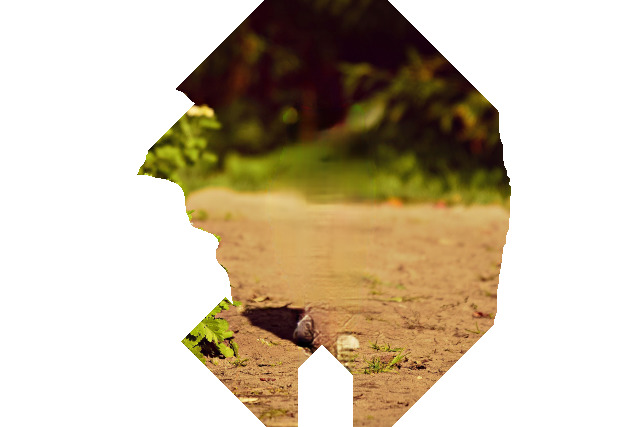}
  {\small\vspace{-4mm}\\w/ dilation}%
}
\vspace{1mm}
\caption{
\tb{Handling imperfect depth edges.} As the detected depth edges may not align well around occlusion boundaries, we dilate the synthesis region by 5 pixels. This strategy helps reduce artifacts in the inpainted regions.
}
\label{fig:erosion_example}
\vspace{-5mm}
\end{figure}
\ignorethis{
\begin{figure}[t]
\fbox{\includegraphics[trim=180 150 180 50,clip,width=\ee]{figure/erosion_example/original}
}%
\hfill%
\fbox{\includegraphics[trim=180 150 180 50,clip,width=\ee]{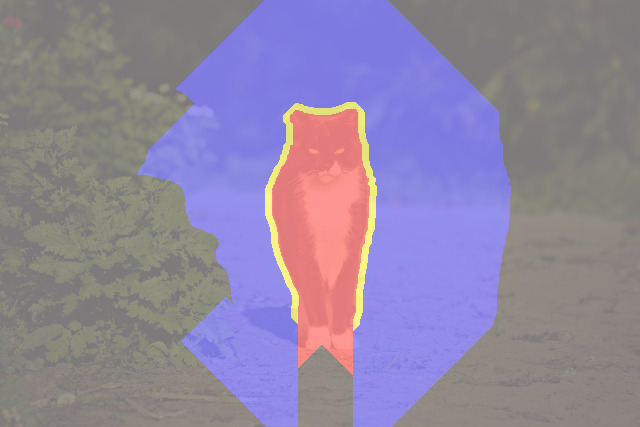}}%
\hfill%
\fbox{\includegraphics[trim=180 150 180 50,clip,width=\ee]{figure/erosion_example/output_wo_erode}}\\

\vspace{\figcapmargin}
\caption{
\tb{Handling imperfect depth edge.} As the depth edges may not align with actual occluding boundary
\jiabin{The EDGE color should match the color in Figure 4 (i.e., BLACK). }
Effect of erosion.
}
\label{fig:erosion_example}
\end{figure}
}
\subsection{Context and synthesis regions}
\label{sec:context_syn_region}

Our inpainting algorithm operates on one of the previously computed depth edges at a time.
Given one of these edges (Figure~\ref{fig:ldi}a), the goal is to synthesize new color and depth content in the adjacent occluded region.
We start by disconnecting the LDI pixels across the discontinuity (Figure~\ref{fig:ldi}b).
We call the pixels that became disconnected (i.e., are now missing a neighbor) \emph{silhouette} pixels.
We see in Figure~\ref{fig:ldi}b that a foreground silhouette (marked green) and a background silhouette (marked red) forms.
Only the background silhouette requires inpainting.
We are interested in extending its surrounding content into the occluded region.

We start by generating a \emph{synthesis region}, a contiguous region of \emph{new} pixels (Figure~\ref{fig:ldi}c, red pixels).
These are essentially just 2D pixel coordinates at this point.
{ 
We initialize the color and depth values in the synthesis region using a simple iterative flood-fill like algorithm.
It starts by stepping from all silhouette pixels one step in the direction where they are disconnected.
These pixels form the initial synthesis region.
We then iteratively expand (for 40 iterations) all pixels of the region by stepping left/right/up/down and adding any pixels that have not been visited before.
For each iteration, we expand the context and synthesis regions alternately and thus a pixel only belong to either one of the two regions
Additionally, we do not step back \emph{across} the silhouette, so the synthesis region remains strictly in the occluded part of the image.
Figure \ref{fig:context_example} shows a few examples.}

We describe our learning-based technique for inpainting the synthesis region in the next section.
Similar techniques \cite{liu2018image,nazeri2019edgeconnect} were previously used for filling holes in images.
One important difference to our work is that these image holes were always fully surrounded by known content, which constrained the synthesis.
In our case, however, the inpainting is performed on a connected layer of an LDI pixels, and it should only be constrained by surrounding pixels that are directly connected to it.
Any other region in the LDI, for example on other foreground or background layer, is entirely irrelevant for this synthesis unit, and should not constrain or influence it in any way.

We achieve this behavior by explicitly defining a \emph{context region} (Figure~\ref{fig:ldi}c, blue region) for the synthesis.
Our inpainting networks only considers the content in the context region and does not see any other parts of the LDI.
The context region is generated using a similar flood-fill like algorithm.
One difference, however, is that this algorithm selects actual LDI pixels and follows their connection links, so the context region expansion halts at silhouettes. 
We run this algorithm for 100 iterations, as we found that synthesis performs better with slightly larger context regions.
In practice, the silhouette pixels may not align well with the actual occluding boundaries due to imperfect depth estimation. To tackle this issue, we dilate the synthesis region near the depth edge by 5 pixels (the context region erodes correspondingly).
%
%
\figref{erosion_example} shows the effect of this heuristic.


\subsection{Context-aware color and depth inpainting}
\label{sec:inpainting}

\heading{Model.} Given the context and synthesis regions, our next goal is to synthesize color and depth values. 
Even though we perform the synthesis on an LDI, the extracted context and synthesis regions are locally like images, so we can use standard network architectures designed for images. 
Specifically, we build our color and depth inpainting models upon image inpainting methods in~\cite{nazeri2019edgeconnect,liu2018image,xiong2019foreground}.

One straightforward approach is to inpaint the color image and depth map independently.
The inpainted depth map, however, may not be well-aligned with respect to the inpainted color.
\begin{figure*}[t]
\centering
\includegraphics[width=0.9\textwidth]{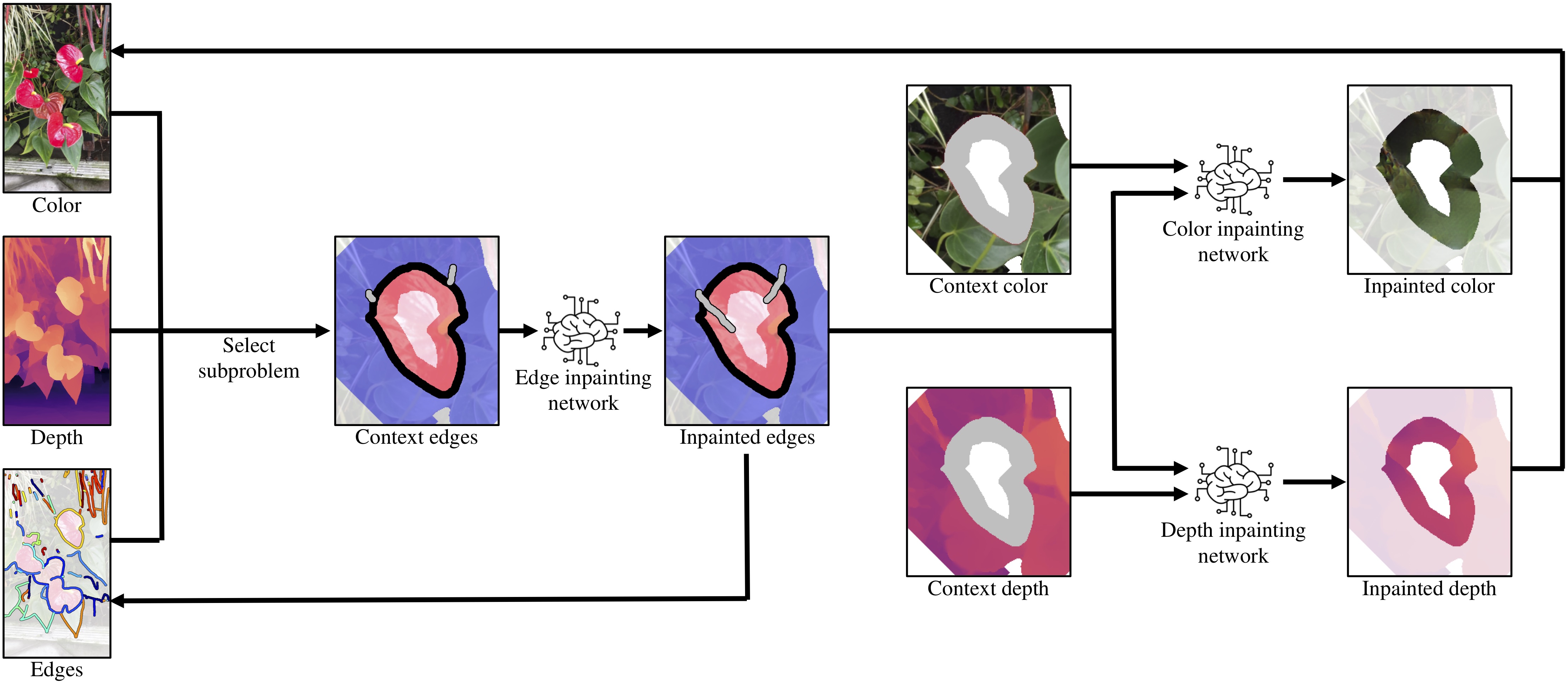}
\caption{\tb{Context-aware color and depth inpainting.} Given the color, depth, the extracted and linked depth edges as inputs, we randomly select one of the edges as a subproblem. 
We start with inpainting the depth edge in the synthesis region (red) using an \emph{edge inpainting network}. 
We then concatenate the inpainted depth edges with the context color together and apply a \emph{color inpainting network} to produce the inpainted color. 
Similarly, we concatenate the inpainted depth edges with the context depth and apply a \emph{depth inpainting network} to produce the inpainted depth. 
}
\vspace{-3mm}
\label{fig:overview}
\end{figure*}

To address this issue, we design our color and depth inpainting network similar to \cite{nazeri2019edgeconnect,xiong2019foreground}:
we break down the inpainting tasks into three sub-networks: (1) edge inpainting network, (2) color inpainting network, and (3) depth inpainting network (\figref{overview}). 
First, given the context edges as input, we use the edge inpainting network to predict the depth edges in the synthesis regions, producing the inpainted edges. 
Performing this step first helps infer the \emph{structure} (in terms of depth edges) that can be used for constraining the \emph{content} prediction (the color and depth values). 
We take the concatenated inpainted edges and context color as input and use the color inpainting network to produce inpainted color. 
We perform the depth inpainting similarly. 
\newlength\deia
\setlength\deia{3.00cm}
\newlength\deib
\setlength\deib{0.007mm}
\newlength\deic
\setlength\deic{1.20cm}
\begin{figure}[h]
\parbox[t]{\deia}{\vspace{0mm}\centering%
  \fbox{\includegraphics[trim=25 0 0 25, clip,width=\deia]%
	    {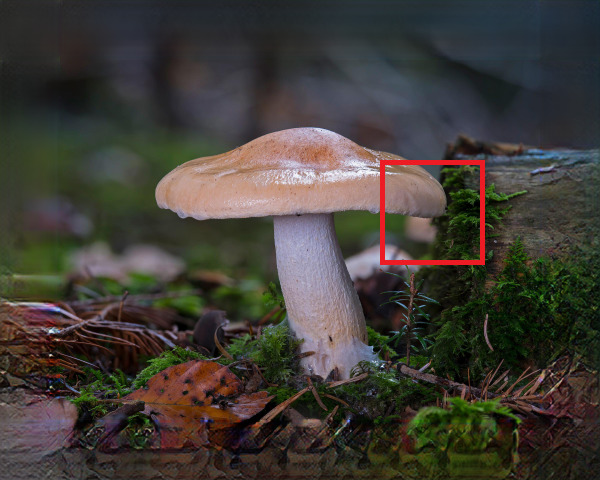}}\vspace{\deib}\\%
}%
\setlength{\fboxrule}{1pt}
\hfill%
\parbox[t]{\deic}{\vspace{0mm}\centering%
  \fcolorbox{red}{red}{\includegraphics[width=\deic]%
	    {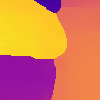}}\\%
  \fcolorbox{red}{red}{\includegraphics[width=\deic]%
	    {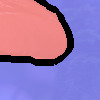}}\\%
    {\footnotesize Zoom-in}\\%
}%
\hfill%
\parbox[t]{\deic}{\vspace{0mm}\centering%
  \fbox{\includegraphics[width=\deic]%
	    {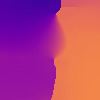}}\\%
 \fcolorbox{blue}{blue}{\includegraphics[width=\deic]%
	    {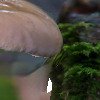}}\\%
	    {\footnotesize Diffusion}\\%
}%
\hfill%
\parbox[t]{\deic}{\vspace{0mm}\centering%
  \fbox{\includegraphics[width=\deic]%
	    {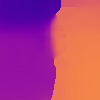}}\\%
 \fcolorbox{blue}{blue}{\includegraphics[width=\deic]%
	    {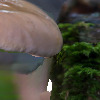}}\\%
	    {\footnotesize w/o edge}\\%
}%
\hfill%
\parbox[t]{\deic}{\vspace{0mm}\centering%
  \fbox{\includegraphics[width=\deic]%
	    {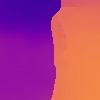}}\\%
  \fcolorbox{blue}{blue}{\includegraphics[width=\deic]%
	    {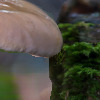}}\\%
	    {\footnotesize w/ edge}\\%
}%
\vspace{0.01mm}%
\caption{\tb{Effect of depth inpainting.} Edge-guided depth inpainting produces more accurate structure inpainting,   particularly for depth-complex regions (e.g., T-junctions). {\color{blue}Blue box}: synthesized novel view.}
\label{fig:depth_edge_inpaint}
\vspace{-0mm}
\end{figure}
\ignorethis{

\begin{figure}[h]
\includegraphics[width=\linewidth]{figure/depth_edge_inpaint/depth_edge_inpaint}
\caption{\tb{Effect of depth inpainting.} Edge-guided depth inpainting produces more accurate structure inpainting,   particularly for depth-complex regions (e.g., T-junctions). 
\jiabin{Please update the image. Inpainted depth and colors under a novel view. Show a) diffusion inpainted depth, b) inpainting WITHOUT edge and c) inpainting WITH edge}
}
\label{fig:depth_edge_inpaint}
\end{figure}
}
\figref{depth_edge_inpaint} shows an example of how the edge-guided inpainting is able to extend the depth structures accurately and alleviate the color/depth misalignment issue.
%

\heading{Multi-layer inpainting.} In depth-complex scenarios, applying our inpainting model once is not sufficient as we can still see the hole through the discontinuity created by the \emph{inpainted} depth edges. 
We thus apply our inpainting model until no further inpainted depth edges are generated. 
\figref{multi_layer} shows an example of the effects. 
Here, applying our inpainting model once fills in missing layers. 
However, several holes are still visible when viewed at a certain viewpoint (\figref{multi_layer}b). 
Applying the inpainting model one more time fixes the artifacts. 
%



\newlength\mlpa
\setlength\mlpa{1.30cm}
\newlength\mlpb
\setlength\mlpb{2.65cm}
\newlength\mlpc
\setlength\mlpc{1.25cm}

\begin{figure}[t]
\parbox[t]{\mlpb}{\vspace{0mm}\centering%
  \includegraphics[width=\mlpb,trim=0 600 400 300,clip]%
	    {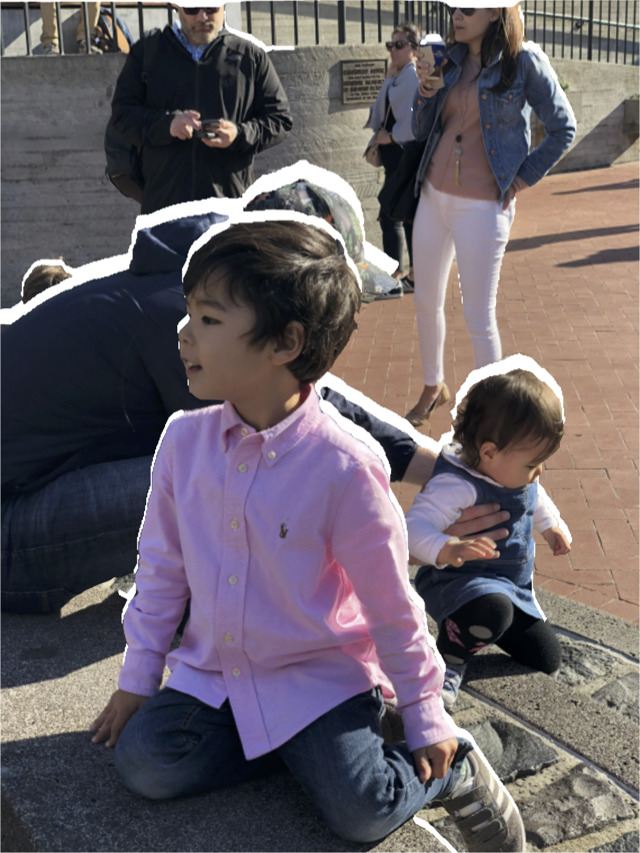}\vspace{0.5mm}\\%
  \fbox{\includegraphics[width=\mlpc]%
	    {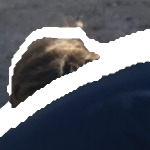}}%
	\hfill%
	\fbox{\includegraphics[width=\mlpc]%
	    {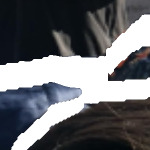}}%
	\vspace{-1mm}\\(a) None%
}%
\hfill%
\parbox[t]{\mlpb}{\vspace{0mm}\centering%
  \includegraphics[width=\mlpb,trim=0 600 400 300,clip]%
	    {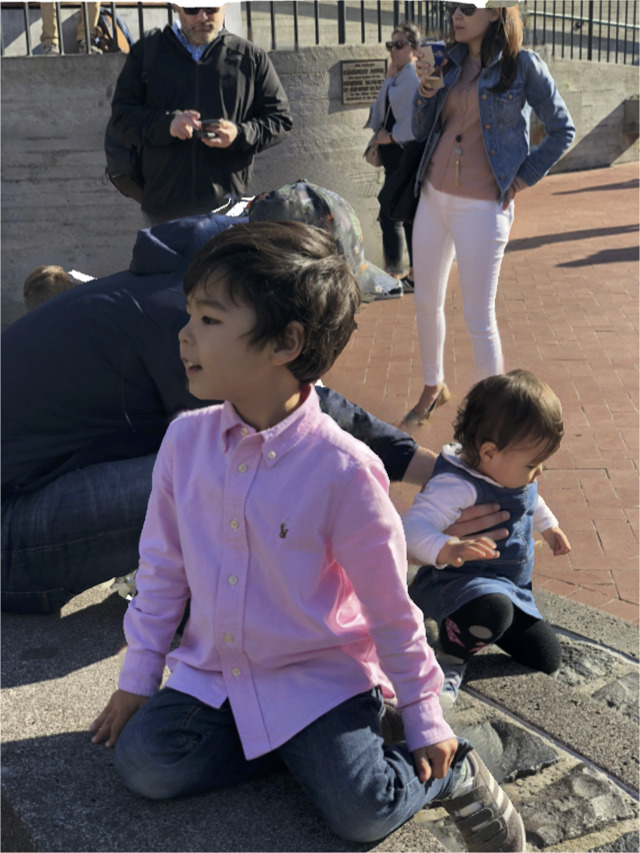}\vspace{0.5mm}\\%
  \fbox{\includegraphics[width=\mlpc]%
	    {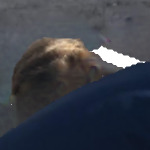}}%
	\hfill%
	\fbox{\includegraphics[width=\mlpc]%
	    {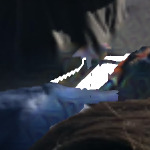}}%
	\vspace{-1mm}\\(b) Once%
}%
\hfill%
\parbox[t]{\mlpb}{\vspace{0mm}\centering%
  \includegraphics[width=\mlpb,trim=0 600 400 300,clip]%
	    {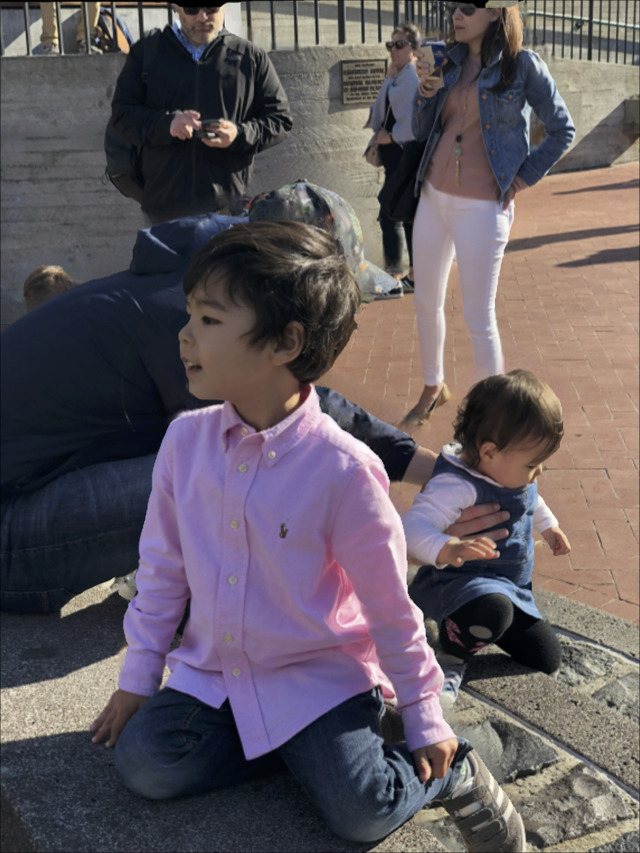}\vspace{0.5mm}\\%
  \fbox{\includegraphics[width=\mlpc]%
	    {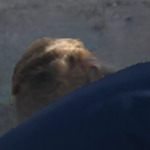}}%
	\hfill%
	\fbox{\includegraphics[width=\mlpc]%
	    {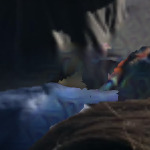}}%
	\vspace{-1mm}\\(c) Twice%
}%
\vspace{\figcapmargin}
\caption{
\tb{Multi-layer inpainting.} 
}
\vspace{-5mm}
\label{fig:multi_layer}
\end{figure}
\heading{Training data generation.} 
For training, our proposed model can be simply trained on any image dataset without the need of annotated data. 
Here, we choose to use MSCOCO dataset~\cite{lin2014microsoft} for its wide diversity in object types and scenes. 
To generate the training data for the inpainting model, we create a synthetic dataset as follows. First, we apply the pre-trained MegaDepth~\cite{li2018megadepth} on the COCO dataset to obtain pseudo ground truth depth maps. 
We extract context/synthesis regions (as described in \secref{context_syn_region}) to form a pool of these regions. 
We then randomly sample and place these context-synthesis regions on \emph{different} images in the COCO dataset. 
We thus can obtain the ground truth content (RGB-D) from the \emph{simulated} occluded region. 

\subsection{Converting to 3D textured mesh}
\label{sec:convert_to_3d}
We form the 3D textured mesh by integrating all the inpainted depth and color values back into the original LDI.
Using mesh representations for rendering allows us to quickly render novel views, without the need to perform per-view inference step. 
Consequently, the 3D representation produced by our algorithm can easily be rendered using standard graphics engines on edge devices.
\section{Experimental Results}
\label{sec:results}
\newlength\qasb
\setlength\qasb{0.35\textwidth}
\newlength\qahl
\setlength\qahl{0.271\qasb}
\newcommand{\boxcolora}{red}
\newcommand{\boxcolorb}{blue}

%

\begin{figure*}
\centering
\parbox[t]{\qasb}{\vspace{0mm}\centering%
  \fbox{\includegraphics[width=\qasb]%
	    {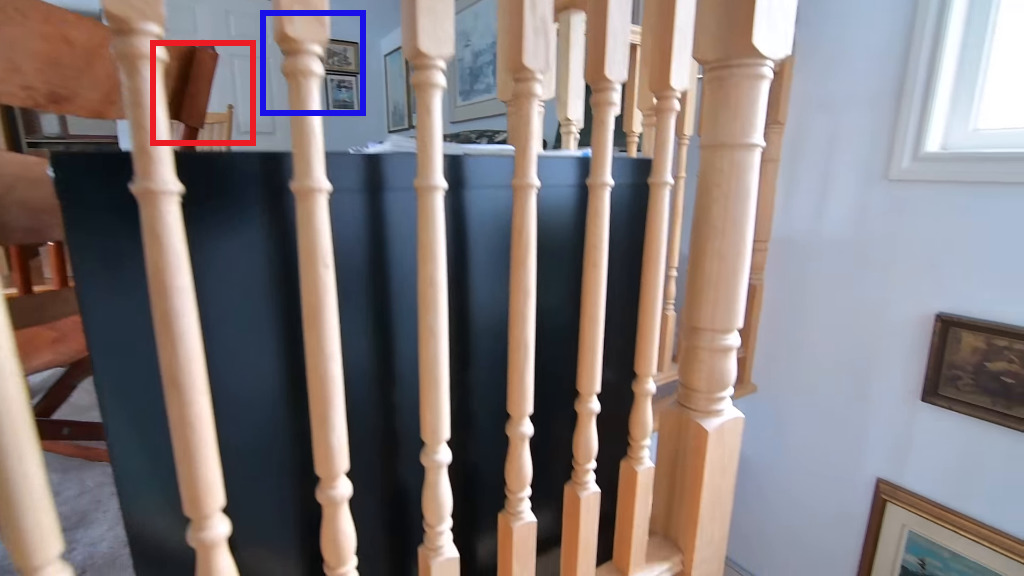}}\vspace{0.5mm}\\%
  \fbox{\includegraphics[width=\qasb]%
	    {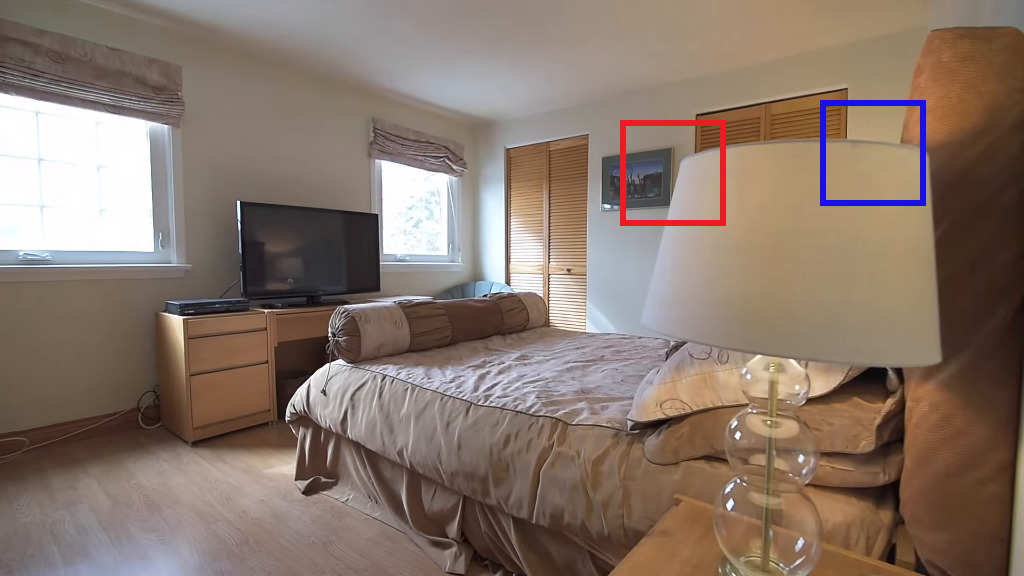}}\vspace{0.5mm}\\
	    Reference Frame\\%
}\hfill%
\setlength{\fboxrule}{1pt}%
\parbox[t]{\qahl}{\vspace{0mm}\centering%
  \fcolorbox{\boxcolora}{\boxcolora}{\includegraphics[width=\qahl]%
	    {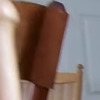}}\\%
  \fcolorbox{\boxcolorb}{\boxcolorb}{\includegraphics[width=\qahl]%
	    {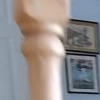}}\vspace{0.5mm}\\%
  \fcolorbox{\boxcolora}{\boxcolora}{\includegraphics[width=\qahl]%
	    {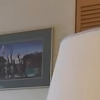}}\\%
  \fcolorbox{\boxcolorb}{\boxcolorb}{\includegraphics[width=\qahl]%
	    {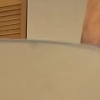}}\\%
	    {\small Zoom-in}\\%
}\hfill%
\parbox[t]{\qahl}{\vspace{0mm}\centering%
  \fcolorbox{\boxcolora}{\boxcolora}{\includegraphics[width=\qahl]%
	    {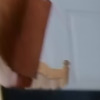}}\\%
  \fcolorbox{\boxcolorb}{\boxcolorb}{\includegraphics[width=\qahl]%
	    {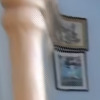}}\vspace{0.5mm}\\%
  \fcolorbox{\boxcolora}{\boxcolora}{\includegraphics[width=\qahl]%
	    {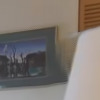}}\\%
  \fcolorbox{\boxcolorb}{\boxcolorb}{\includegraphics[width=\qahl]%
	    {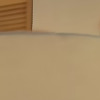}}\\%
	    {\footnotesize StereoMag}{\small~\cite{zhou2018stereo}}\\%
}\hfill%
\parbox[t]{\qahl}{\vspace{0mm}\centering%
  \fcolorbox{\boxcolora}{\boxcolora}{\includegraphics[width=\qahl]%
	    {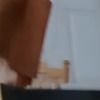}}\\%
  \fcolorbox{\boxcolorb}{\boxcolorb}{\includegraphics[width=\qahl]%
	    {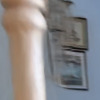}}\vspace{0.5mm}\\%
  \fcolorbox{\boxcolora}{\boxcolora}{\includegraphics[width=\qahl]%
	    {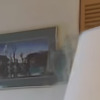}}\\%
  \fcolorbox{\boxcolorb}{\boxcolorb}{\includegraphics[width=\qahl]%
	    {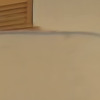}}\\%
	 {\small PB-MPI~\cite{srinivasan2019pushing}}\\%
}\hfill%
\parbox[t]{\qahl}{\vspace{0mm}\centering%
  \fcolorbox{\boxcolora}{\boxcolora}{\includegraphics[width=\qahl]%
	    {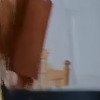}}\\%
  \fcolorbox{\boxcolorb}{\boxcolorb}{\includegraphics[width=\qahl]%
	    {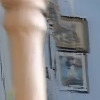}}\vspace{0.5mm}\\%
  \fcolorbox{\boxcolora}{\boxcolora}{\includegraphics[width=\qahl]%
	    {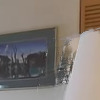}}\\%
  \fcolorbox{\boxcolorb}{\boxcolorb}{\includegraphics[width=\qahl]%
	    {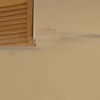}}\\%
	 {\small LLFF~\cite{mildenhall2019llff}}\\%
}\hfill%
\parbox[t]{\qahl}{\vspace{0mm}\centering%
  \fcolorbox{\boxcolora}{\boxcolora}{\includegraphics[width=\qahl]%
	    {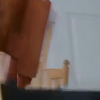}}\\%
  \fcolorbox{\boxcolorb}{\boxcolorb}{\includegraphics[width=\qahl]%
	    {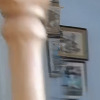}}\vspace{0.5mm}\\%
  \fcolorbox{\boxcolora}{\boxcolora}{\includegraphics[width=\qahl]%
	    {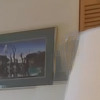}}\\%
  \fcolorbox{\boxcolorb}{\boxcolorb}{\includegraphics[width=\qahl]%
	    {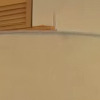}}\\%
	 {\small XView~\cite{choi2019extreme}}\\%
}\hfill%
\parbox[t]{\qahl}{\vspace{0mm}\centering%
  \fcolorbox{\boxcolora}{\boxcolora}{\includegraphics[width=\qahl]%
	    {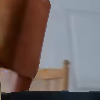}}\\%
  \fcolorbox{\boxcolorb}{\boxcolorb}{\includegraphics[width=\qahl]%
	    {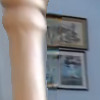}}\vspace{0.5mm}\\%
  \fcolorbox{\boxcolora}{\boxcolora}{\includegraphics[width=\qahl]%
	    {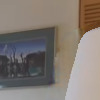}}\\%
  \fcolorbox{\boxcolorb}{\boxcolorb}{\includegraphics[width=\qahl]%
	    {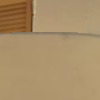}}\\%
  {\small Ours}\\%
}\hfill%
\vspace{\figcapmargin}
\caption{\tb{Visual comparison with MPI-based methods.} Our method inpaints plausible structure and color in the occluded region.
}
\label{fig:visual_mpi}
\end{figure*}

\newlength\qfba
\setlength\qfba{4.40cm}
\newlength\qfbb
\setlength\qfbb{2.16cm}
\newcommand\at{0.5}
\def\qfaa{0.75}
\def\qfab{0.75920245398}
\def\qfac{0.7941176470588235}

\begin{figure*}[t]
\centering%
\parbox[t]{\qfaa\qfba}{\vspace{0mm}\centering%
  \includegraphics[width=\qfaa\qfba,height=\qfba]%
	    {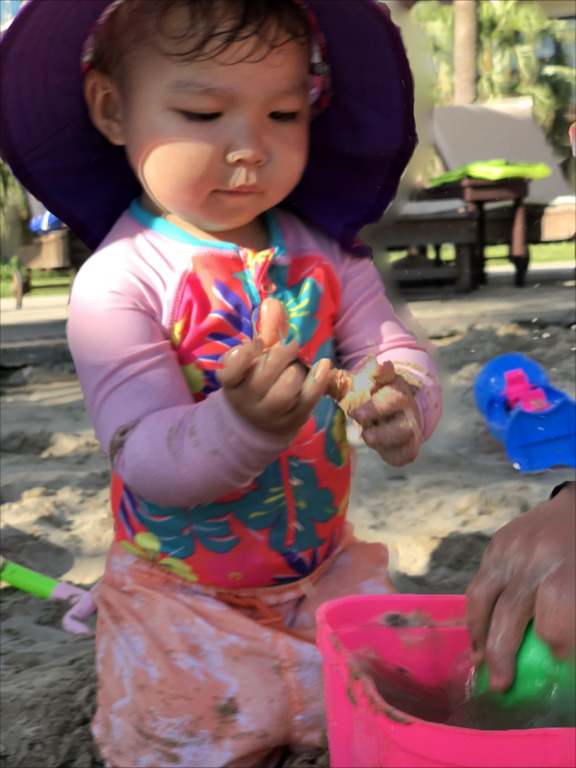}}%
\hspace{1mm}%
\parbox[t]{\qfbb}{\vspace{0mm}\centering%
  \includegraphics[width=\qfbb,height=\qfbb]%
	    {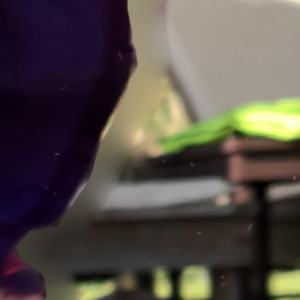}\vspace{0.5mm}\\%
  \includegraphics[width=\qfbb,height=\qfbb]%
	    {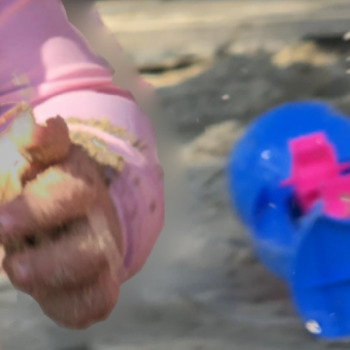}}%
%
\hfill%
\parbox[t]{\qfab\qfba}{\vspace{0mm}\centering%
  \includegraphics[width=\qfab\qfba,height=\qfba]%
	    {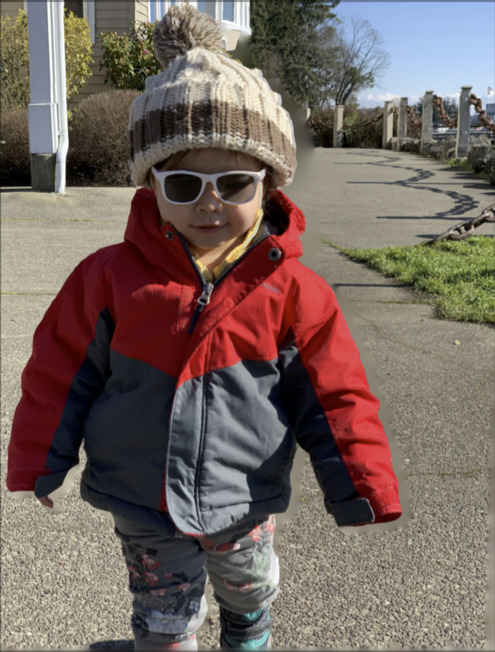}}%
\hspace{1mm}%
\parbox[t]{\qfbb}{\vspace{0mm}\centering%
  \includegraphics[width=\qfbb,height=\qfbb]%
	    {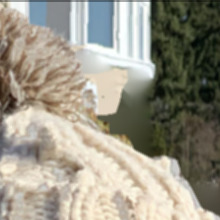}\vspace{0.5mm}\\%
  \includegraphics[width=\qfbb,height=\qfbb]%
	    {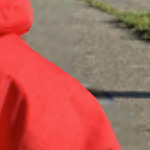}}%
%
\hfill%
\parbox[t]{\qfac\qfba}{\vspace{0mm}\centering%
\includegraphics[width=\qfac\qfba,height=\qfba]%
	    {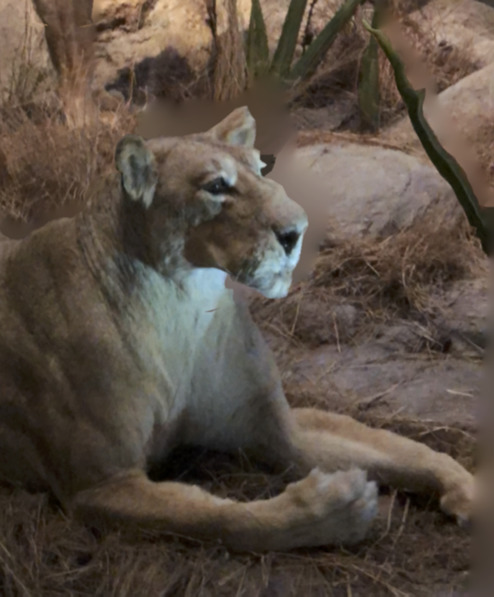}}%
\hspace{1mm}%
\parbox[t]{\qfbb}{\vspace{0mm}\centering%
\includegraphics[width=\qfbb,height=\qfbb]%
	    {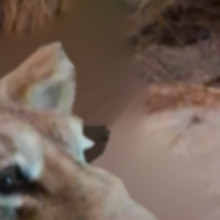}\vspace{0.5mm}\\%
  \includegraphics[width=\qfbb,height=\qfbb]%
	    {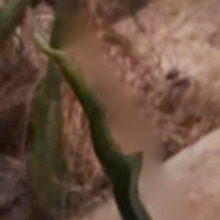}}%
%
\vspace{0mm}\\%
$\underbracket[1pt][2.0mm]{\hspace{\textwidth}}_%
    {\substack{\vspace{-3.0mm}\\\colorbox{white}{~~Facebook 3D Photo results~~}}}$\vspace{1mm}\\%
\parbox[t]{\qfaa\qfba}{\vspace{0mm}\centering%
  \includegraphics[width=\qfaa\qfba,height=\qfba]%
	    {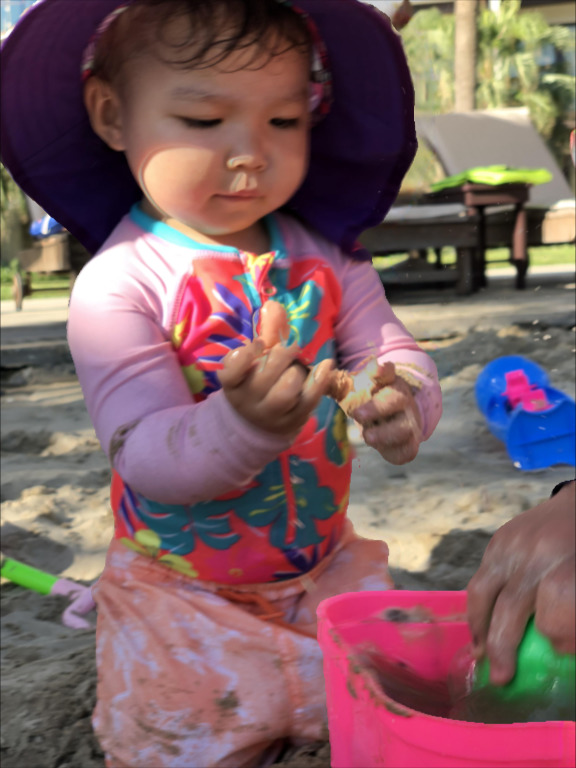}}%
\hspace{1mm}%
\parbox[t]{\qfbb}{\vspace{0mm}\centering%
  \includegraphics[width=\qfbb,height=\qfbb]%
	    {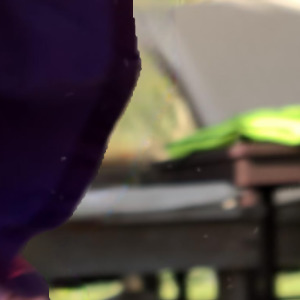}\vspace{0.5mm}\\%
  \includegraphics[width=\qfbb,height=\qfbb]%
	    {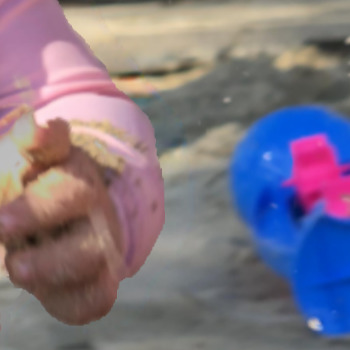}}%
%
\hfill%
\parbox[t]{\qfab\qfba}{\vspace{0mm}\centering%
  \includegraphics[width=\qfab\qfba,height=\qfba]%
	    {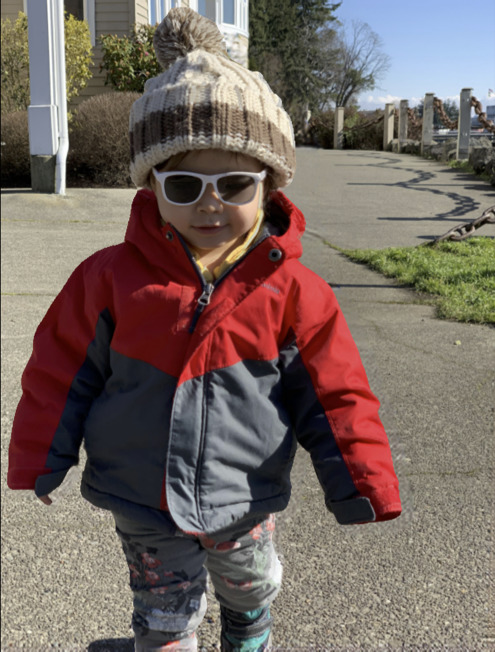}}%
\hspace{1mm}%
\parbox[t]{\qfbb}{\vspace{0mm}\centering%
  \includegraphics[width=\qfbb,height=\qfbb]%
	    {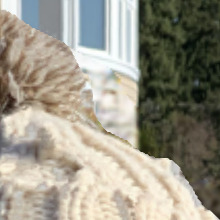}\vspace{0.5mm}\\%
  \includegraphics[width=\qfbb,height=\qfbb]%
	    {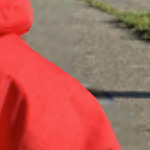}}%
%
\hfill%
\parbox[t]{\qfac\qfba}{\vspace{0mm}\centering%
  \includegraphics[width=\qfac\qfba,height=\qfba]%
	    {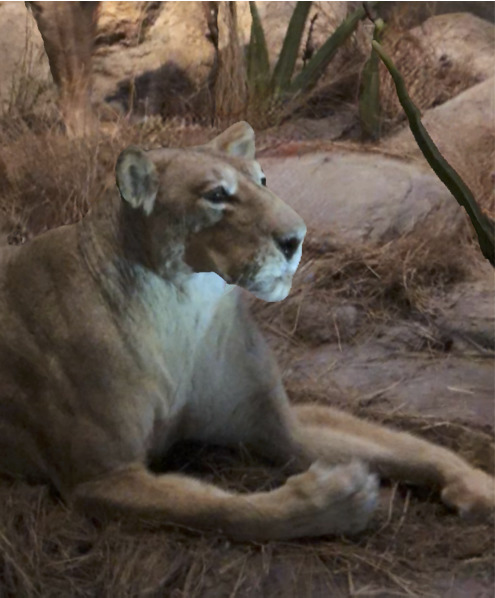}}%
\hspace{1mm}%
\parbox[t]{\qfbb}{\vspace{0mm}\centering%
  \includegraphics[width=\qfbb,height=\qfbb]%
	    {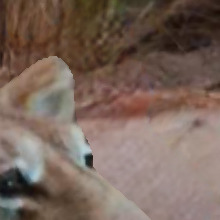}\vspace{0.5mm}\\%
 \includegraphics[width=\qfbb,height=\qfbb]%
	    {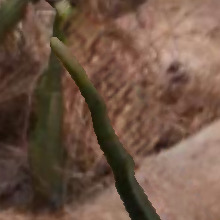}}%
%
%
\vspace{0mm}\\%
$\underbracket[1pt][2.0mm]{\hspace{\textwidth}}_%
    {\substack{\vspace{-3.0mm}\\\colorbox{white}{~~Our results~~}}}$\vspace{1mm}\\%
\caption{\tb{Visual comparison to Facebook 3D Photos.} Our approach fills plausible textures and structures at disocclusions.}
\vspace{-3mm}
\label{fig:qualitative_fb3d}
\end{figure*}
In this section, we start with describing implementation details (\secref{details}).
We then show visual comparisons with the state-of-the-art novel view synthesis methods (\secref{visual}).
We refer to the readers to supplementary material for extensive results and comparisons.
Next, we follow the evaluation protocol in ~\cite{zhou2018stereo} and report the quantitative comparisons on the RealEstate10K dataset (\secref{quantitative}).
We present an ablation study to justify our model design (\secref{ablation}).
Finally, we show that our method works well with depth maps from different sources (\secref{depth_quality}).
Additional details and visual comparisons can be found in our supplementary material.

\subsection{Implementation details}\label{sec:details}
\heading{Training the inpainting model.} 
For the edge-generator, we follow the hyper-parameters in~\cite{nazeri2019edgeconnect}. 
Specifically, we train the edge-generator model using the ADAM optimizer~\cite{kingma2014adam} with $\beta=0.9$ and an initial learning rate of $0.0001$. 
We train both the edge and depth generator model using the context-synthesis regions dataset on the MS-COCO dataset for 5 epochs. 
We train the depth generator and color image generator for 5 and 10 epochs, respectively.

\heading{Inpainting model architecture.} 
For the edge inpainting network, we adopt the architecture provided by~\cite{nazeri2019edgeconnect}. 
For the depth and color inpainting networks, we use a standard U-Net architecture with partial covolution~\cite{liu2018image}. 
Due to the space limitation, we leave additional implementation details (specific network architecture, the training loss and the weights for each network) to the supplementary material. 
We will make the source code and pre-trained model publicly available to foster future work.

\heading{Training data.} We use the 118k images from COCO 2017 set for training. We select at most 3 pairs of regions from each image to form the context-synthesis pool. During training, we sample one pair of regions for each image, and resize it by a factor between $\left[1.0, 1.3\right]$.


\subsection{Visual comparisons}
\label{sec:visual}
\heading{Comparisons with methods with MPI representations.} 
We compare our proposed model against MPI-based approaches on RealEstate10K dataset. 
We use DPSNet~\cite{im2019dpsnet} to obtain the input depth maps for our method.
We render the novel views of MPI-based methods using the pre-trained weights provided by the authors. 
\figref{visual_mpi} shows two challenging examples with complex depth structures. Our method synthesizes plausible structures around depth boundaries; on the other hand, stereo magnification and PB-MPI produce artifacts around depth discontinuities. 
LLFF~\cite{mcmillan1995plenoptic} suffers from ghosting effects when extrapolating new views.
\ignorethis{
Stereo magnification takes a stereo pair of images as input. 
Here, we use the examples provided by~\cite{zhou2018stereo} but use only the right color image. We obtain the corresponding depth maps of these images using MegaDepth~\cite{li2018megadepth}. We use the aligned color/depth images as inputs for our algorithm. To obtain the results of stereo magnification, we use the pre-trained model to produce multi-plane images for each testing ex\-am\-ple and use them to render the scene at novel viewpoints\footnote{Available at \url{https://github.com/google/stereo-magnification}}.
\figref{qualitative_mpi} shows two examples of challenging scenes.  While the stereo magnification method produces plausible view extrapolation results along the camera baseline direction, the synthesized novel views reveal clearly visible artifacts, particularly along the object boundaries. 
}

\heading{Comparisons with Facebook 3D photo.} 
Here, we aim to evaluate the capability of our method on photos taken \emph{in the wild}. 
We extract the color images and the corresponding depth maps estimated from an iPhone X (with dual camera lens). 
We use the same set of RGB-D inputs for both Facebook 3D photo and our algorithm.
\figref{qualitative_fb3d} shows the view synthesis result in comparison with Facebook 3D photo. 
The diffused color and depth values by the facebook 3D photo algorithm work well when small or thin occluded regions are revealed at novel views. 
These artifacts, however, become clearly visible with larger occluded regions. 
On the other hand, our results in general fills in the synthesis regions with visually plausible contents and structures.

\ignorethis{
\newlength\qlsa
\setlength\qlsa{4.10cm}
\newlength\qlsb
\setlength\qlsb{1.0mm}

\def\qlaa{0.3153988868274583}
\def\qlab{0.2565217391304348}
\def\qlac{0.3196969696969697}

\begin{figure}%
\centering%
\parbox[t]{\qlsa}{\vspace{0mm}\centering%
  \includegraphics[width=\qlsa,height=\qlaa\qlsa]%
	    {figure/qualitative_lsi/LSI_ex1_baseline}\vspace{\qlsb}\\%
  \includegraphics[width=\qlsa,height=\qlab\qlsa]%
	    {figure/qualitative_lsi/LSI_ex2_baseline}\vspace{\qlsb}\\%
  \includegraphics[width=\qlsa,height=\qlac\qlsa]%
	    {figure/qualitative_lsi/LSI_ex4_baseline}\vspace{-1mm}\\%
	LSI results~\cite{tulsiani2018layer}}%
\hfill%
\parbox[t]{\qlsa}{\vspace{0mm}\centering%
  \includegraphics[width=\qlsa,height=\qlaa\qlsa]%
	    {figure/qualitative_lsi/LSI_ex1_ours}\vspace{\qlsb}\\%
  \includegraphics[width=\qlsa,height=\qlab\qlsa]%
	    {figure/qualitative_lsi/LSI_ex2_ours}\vspace{\qlsb}\\%
  \includegraphics[width=\qlsa,height=\qlac\qlsa]%
	    {figure/qualitative_lsi/LSI_ex4_ours}\vspace{-1mm}\\%
	Our results}\vspace{1mm}\\%
\caption{Qualitative results compare to LSI~\cite{tulsiani2018layer}.}
\label{fig:qualitative_lsi}

\end{figure}
\heading{Comparisons with Layered Scene Inference~\cite{tulsiani2018layer}.} 
As our method shares similar 3D representation as \cite{tulsiani2018layer} (layered depth images), we show visual comparison to highlight the importance of performing the proposed localized inpainting with spatially adaptive contexts. 
As the model from LSI~\cite{tulsiani2018layer} is \emph{domain-specific}, we evaluate the pre-trained model provided by the authors on images from the KITTI raw dataset. 
\figref{qualitative_lsi} shows visual comparisons with three examples. 
While our model is \emph{domain-agnostic} (e.g., our inpainting model has \emph{not} trained on street view images), our results contain significantly fewer artifacts.
}

\begin{table}[t]
\caption{\tb{Quantitative comparison} on the RealEstate10K dataset.}
\label{tab:quantitative_other}
\centering
\small
\begin{tabular}{l|ccc}
\toprule
    Methods &   SSIM $\uparrow$    &   PSNR $\uparrow$   &   LPIPS $\downarrow$    \\ 
    \midrule
    Stereo-Mag~\cite{zhou2018stereo}   &  \tb{0.8906} & 	26.71 &	0.0826   \\ 
    PB-MPI~\cite{srinivasan2019pushing}   &     0.8773    &   25.51   & 0.0902   \\ 
    LLFF~\cite{mildenhall2019llff}   &   0.8697   &   24.15   &   0.0941   \\ 
    Xview~\cite{choi2019extreme}   &   0.8628   &   24.75   &   0.0822   \\ 
    Ours   &   0.8887       &   \tb{27.29}  &  \tb{0.0724}     \\ 
    \bottomrule
\end{tabular}
\end{table}

\subsection{Quantitative comparisons}
\label{sec:quantitative}
We evaluate how well our model can extrapolate views compared to MPI-based methods~\cite{srinivasan2019pushing,zhou2018stereo,choi2019extreme,mildenhall2019llff}. 
%
We randomly sample 1500 video sequences from RealEstate10K to generate testing triplets. 
For each triplet, we set $t=10$ for target view, so that all the methods need to extrapolate beyond the source ($t=0$) and reference ($t=4$) frame. 
We use DPSNet~\cite{im2019dpsnet} to generate the input depth maps required for our model.
We quantify the performance of each model using SSIM and PSNR metrics between the synthesized target views and the ground truth. 
As these metrics do not capture the perceptual quality of the synthesized view, we include LPIPS~\cite{zhang2018unreasonable} metric to quantify how well does the generated view align with human perception. 
For PB-MPI, we set the number of depth layers to 64 as it yields the best result.
We report the evaluation results in~\tabref{quantitative_other}. 
Our proposed method performs competitively on SSIM and PSNR. 
In addition, our synthesis views exhibit better perceptual quality, as reflected in the superior LPIPS score. 
\subsection{Ablation study}
\label{sec:ablation}
\newcommand{\qn}[1]{(\blue{#1})}
\begin{table}[t]
\caption{\tb{ Using depth edge as guidance improves the results.} \blue{Blue}: results in disocculded regions.}
\label{tab:ablation}
\centering
\setlength{\tabcolsep}{3pt}
\small
\begin{tabular}{l|ccc}
\toprule
    Methods &   SSIM $\uparrow$    &   PSNR $\uparrow$   &   LPIPS $\downarrow$    \\ 
\midrule
    Diffusion   &   0.8665 \qn{0.6237}   &   25.95 \qn{18.91}   &   0.084   \\ 
    Inpaint w/o edge   &   0.8665 \qn{0.6247} &   25.96 \qn{18.94}   &   0.084   \\ 
    Inpaint w/ edge (Ours)   &   0.8666 \qn{0.6265}   &   25.97 \qn{18.98}   &   0.083   \\ 
    \bottomrule
\end{tabular}
\end{table}

\begin{table}[t]
\caption{\tb{Using color inpainting model gives better perceptual quality}. Our dilation heuristic further boosts the performance. \blue{Blue}: results in disocculded regions.}%
\label{tab:ablation2}
\centering
\small
\setlength{\tabcolsep}{2pt}
\begin{tabular}{l|ccc}
\toprule
    Methods &   SSIM $\uparrow$    &   PSNR $\uparrow$   &   LPIPS $\downarrow$    \\ 
\midrule
    Diffusion   &   0.8661 \qn{0.6215}  &   25.90 \qn{18.78}  &   0.088   \\ 
    Inpaint w/o dilation   &   0.8643 \qn{0.5573}  &   25.56 \qn{17.14}  &   0.085   \\ 
    Inpaint w/  dilation (Ours)   &   0.8666 \qn{0.6265}  &   25.97 \qn{18.98}  &   0.083   \\ 
    \bottomrule
\end{tabular}
\end{table}
\newlength\cia
\setlength\cia{0.49\columnwidth}
\newlength\cib
\setlength\cib{0.25\columnwidth}
\begin{figure}[h]
\setlength{\fboxrule}{1pt}%
\parbox[t]{\cia}{\vspace{0mm}\centering%
  \includegraphics[trim=224 0 224 0,clip,width=\cia,height=1.95\cib]%
	    {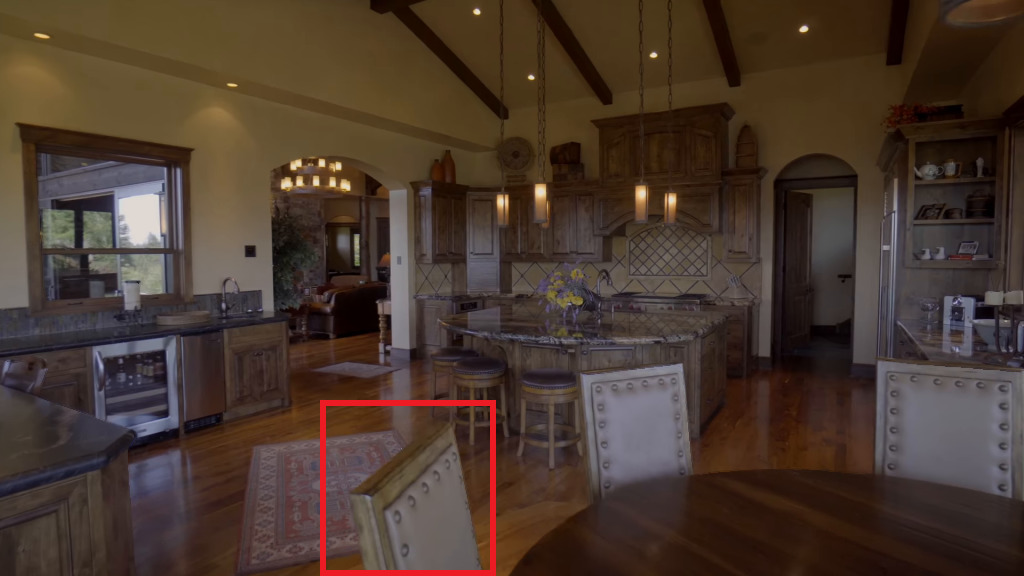} \\
	    {\vspace{-1mm}\footnotesize Input}\\%
}%
\hfill%
\parbox[t]{\cib}{\vspace{0mm}\centering%
  \includegraphics[width=\cib]%
	    {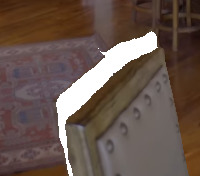}\\%
  {\vspace{-1mm}\footnotesize(a) Disocclusion}\\
  \includegraphics[width=\cib]%
	    {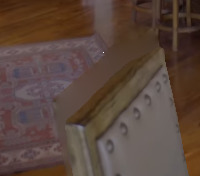}\\%
  {\vspace{-1mm}\footnotesize(c) w/o Dilation}
}%
\hfill%
\parbox[t]{\cib}{\vspace{0mm}\centering%
  \includegraphics[width=\cib]%
	    {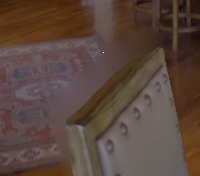}\\%
  {\vspace{-1mm}\footnotesize(b) Diffusion}\\
  \includegraphics[width=\cib]%
	    {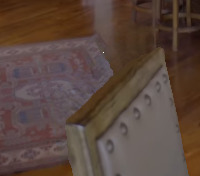}\\%
  {\vspace{-1mm}\footnotesize(d) w/ Dilation}
}%
\caption{\tb{Color inpainting leads to better visual quality.}
}
\label{fig:color_inpaint}
\vspace{-1mm}
\end{figure}

We conduct ablation studies to see how each of our proposed components contribute to the final performance. 
We first verify the effectiveness of \emph{edge-guided depth inpainting}. 
We sample 130 triplets from our testing sequences, evaluate the {inpainted color on both the entire image and disoccluded regions}, and report the numbers in~\tabref{ablation}. 
The results show that our proposed edge-guided inpainting leads to minor improvement in numerical metrics.
Next, we examine the efficacy of our \emph{color inpainting} model following the same procedure described above. 
We present the performance in both entire image and occluded regions in~\tabref{ablation2}. 
We observe that our proposed model yields better perceptual quality.
\figref{color_inpaint} shows an example.

\newlength\ada
\setlength\ada{2.00cm}
\newlength\adb
\setlength\adb{0.5mm}

\begin{figure}%
\centering%
\parbox[t]{\ada}{\vspace{0mm}\centering%
  \includegraphics[width=\ada]%
	    {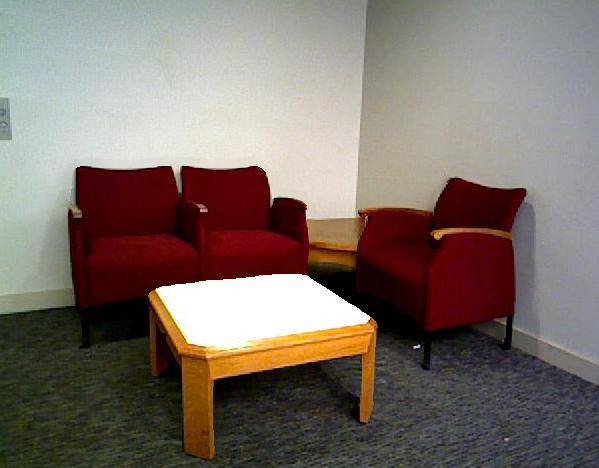}\vspace{\adb}\\%
  \includegraphics[width=\ada]%
	    {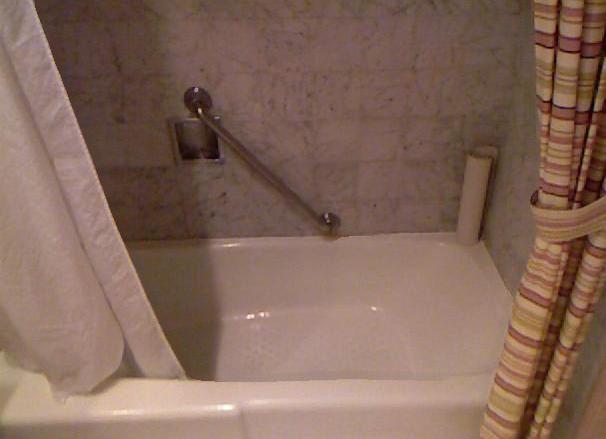}\vspace{\adb}\\%
	Input}%
\hfill%
\parbox[t]{\ada}{\vspace{0mm}\centering%
  \includegraphics[width=\ada]%
	    {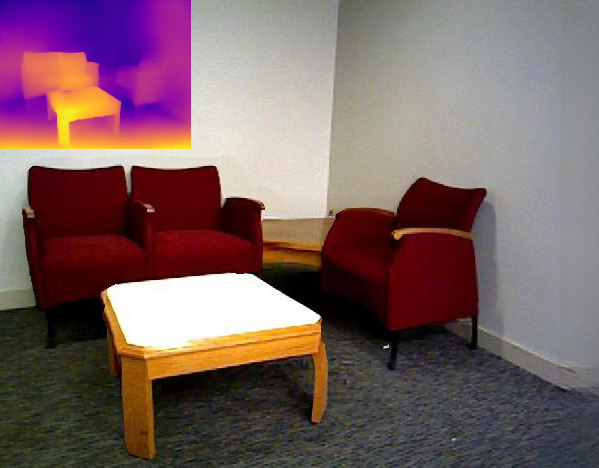}\vspace{\adb}\\%
  \includegraphics[width=\ada]%
	    {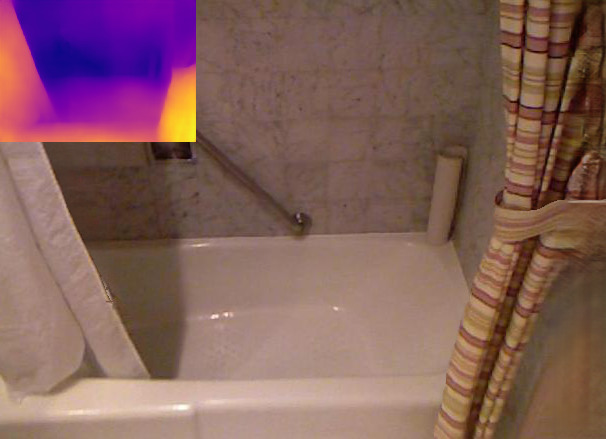}\vspace{\adb}\\%
	MegaDepth}%
\hfill%
\parbox[t]{\ada}{\vspace{0mm}\centering%
  \includegraphics[width=\ada]%
	    {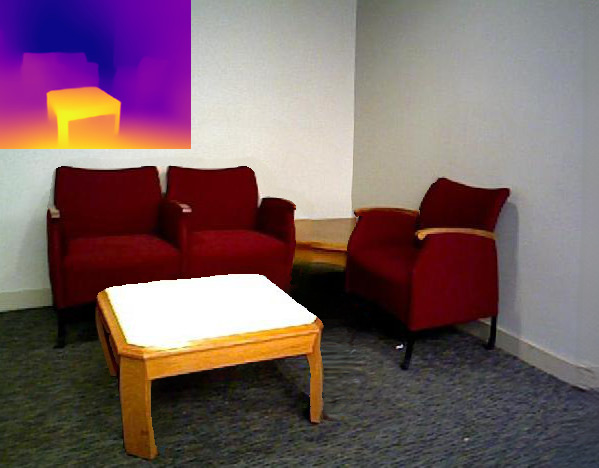}\vspace{\adb}\\%
  \includegraphics[width=\ada]%
	    {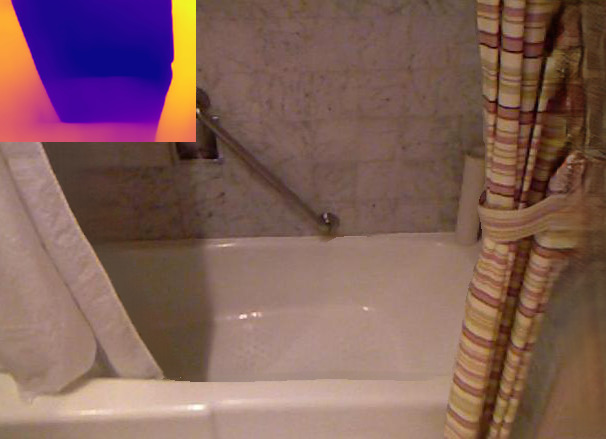}\vspace{\adb}\\%
	MiDas}%
\hfill%
\parbox[t]{\ada}{\vspace{0mm}\centering%
  \includegraphics[width=\ada]%
	    {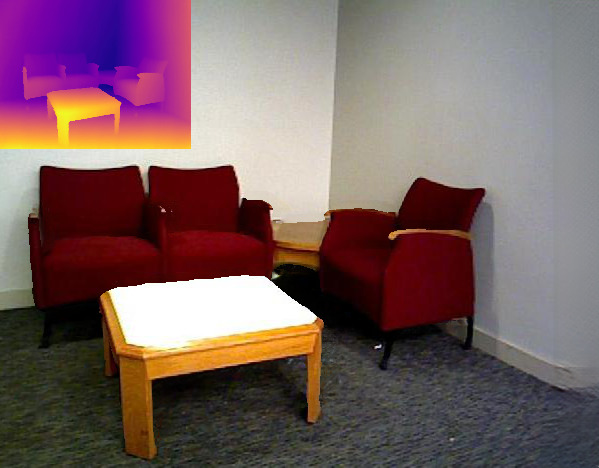}\vspace{\adb}\\%
  \includegraphics[width=\ada]%
	    {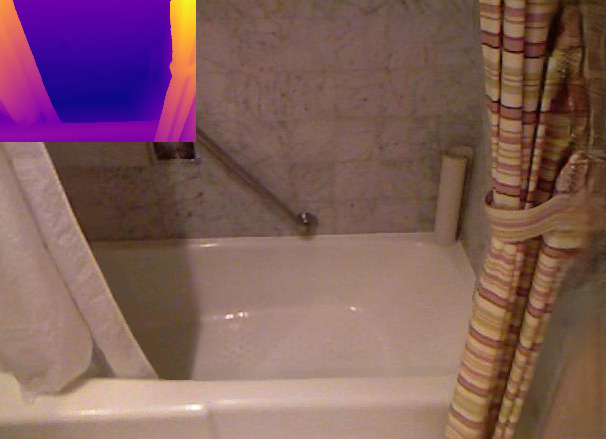}\vspace{\adb}\\%
	Kinect}%
\vspace{\adb}
\caption{\tb{Our method works with various sources of depth map.} We show the depth estimates on the top-left of novel views.}
\label{fig:different_depth}
\vspace{-5mm}
\end{figure}

\subsection{Handling different depth maps}
\label{sec:depth_quality}
We test our method using depth maps generated using different approaches (\figref{different_depth}). We select images from SUNRGBD~\cite{song2015sun} dataset, and obtain the corresponding depth maps from three different sources: 1) depth estimated with MegaDepth~\cite{li2018megadepth}, 2) MiDas~\cite{lasinger2019towards} and 3) Kinect depth sensor. We present the resulting 3D photos in \figref{different_depth}. 
The results show that our method can handle depth maps from different sources reasonably well.

\ignorethis{
\subsection{Historical photos.} 
\label{sec:historical}
Here we apply our method to several historical photos. 
We extract the depth map using the pre-trained MiCaS model~\cite{lasinger2019towards}. 
\figref{fig_historical_photo} shows the input images and the rendered novel views. 
This highlights the wide applicability of our approach.
\newlength\hpa
\setlength\hpa{2.7cm}
\newlength\hpb
\setlength\hpb{0.001mm}

\begin{figure}[t]
\parbox[t]{\hpa}{\centering%
  \includegraphics[width=\hpa]{figure/historical_photo/example_1/churchill_ori}
\includegraphics[width=\hpa]{figure/historical_photo/example_2/moon_ori}\vspace{\hpb}\\%
	Input}%
\hfill%
\parbox[t]{\hpa}{\centering%
  \includegraphics[width=\hpa]{figure/historical_photo/example_1/churchill_mask}
\includegraphics[width=\hpa]{figure/historical_photo/example_2/moon_mask}\vspace{\hpb}\\%
	Synthesis region}%
\hfill%
\parbox[t]{\hpa}{\centering%
  \includegraphics[width=\hpa]{figure/historical_photo/example_1/churchill_pred}
\includegraphics[width=\hpa]{figure/historical_photo/example_2/moon_pred}\vspace{\hpb}\\%
	Novel view}%
\vspace{\figcapmargin}
\caption{\textbf{3D Photos of historical pictures.}}
\label{fig:fig_historical_photo}
\end{figure}

\ignorethis{
\begin{figure}[h]
\includegraphics[width=\linewidth]{figure/historical_photo/historical_photo}
\label{fig:fig_historical_photo}
\caption{3D Photo from historical photo.}
\end{figure}
}

}

\ignorethis{
\newlength\fasb
\setlength\fasb{5.75cm}
\newlength\fasc
\setlength\fasc{1.59cm}
\newlength\fasep
\setlength\fasep{-7px}
\newlength\fasd
\setlength\fasd{2.025\fasc}
\begin{figure*}[t]%
\parbox[t]{\fasb}{\vspace{0mm}\centering%
  \fbox{\includegraphics[width=\fasb,trim=0 0 0 0,clip]%
	    {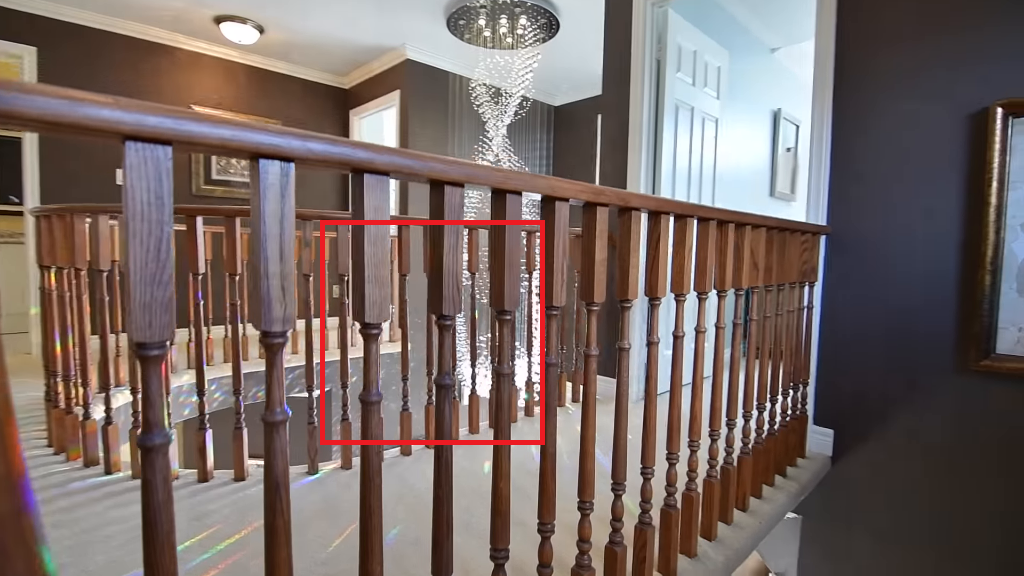}}\vspace{1mm}\\%
  \fbox{\includegraphics[width=\fasb,trim=0 0 0 0,clip]%
	    {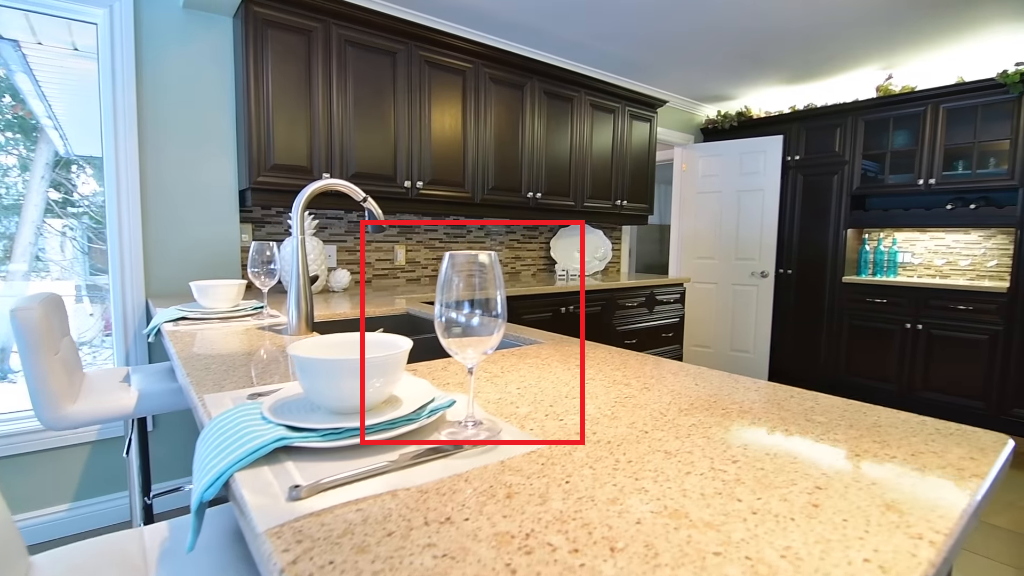}\vspace{1mm}}\\%
}\hfill%
\parbox[t]{\fasc}{\vspace{0mm}\centering%
\setlength{\fboxrule}{1pt}
  \fcolorbox{red}{red}{\includegraphics[width=\fasc,trim=0 0 0 0,clip]%
	    {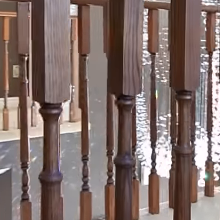}}\vspace{-1mm}\\%
  \fcolorbox{red}{red}{\includegraphics[width=\fasc,trim=0 0 0 0,clip]%
	    {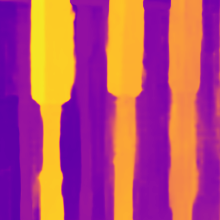}}\vspace{1mm}\\%
  \fcolorbox{red}{red}{\includegraphics[width=\fasc,trim=0 0 0 0,clip]%
	    {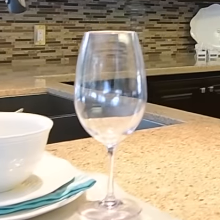}}\vspace{-1mm}\\%
  \fcolorbox{red}{red}{\includegraphics[width=\fasc,trim=0 0 0 0,clip]%
	    {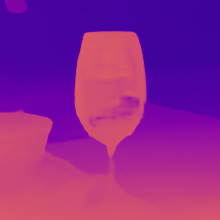}\vspace{-1mm}}\\%
}\hfill%
\parbox[t]{\fasd}{\vspace{0mm}\centering%
  \fbox{\includegraphics[width=\fasd,trim=0 0 0 0,clip]%
	    {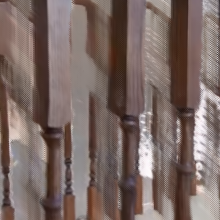}}\vspace{1mm}\\%
  \fbox{\includegraphics[width=\fasd,trim=0 0 0 0,clip]%
	    {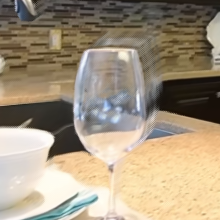}}\vspace{-1mm}\\%
  {StereoMag~\cite{zhou2018stereo}}
}\hfill%
\parbox[t]{\fasd}{\vspace{0mm}\centering%
  \fbox{\includegraphics[width=\fasd,trim=0 0 0 0,clip]%
	    {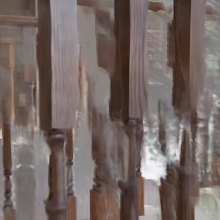}}\vspace{1mm}\\%
  \fbox{\includegraphics[width=\fasd,trim=0 0 0 0,clip]%
	    {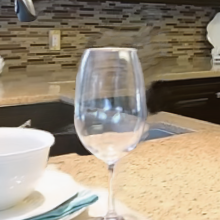}}\vspace{-1mm}\\%
  {PB-MPI~\cite{srinivasan2019pushing}}  
}\hfill%
\parbox[t]{\fasd}{\vspace{0mm}\centering%
  \fbox{\includegraphics[width=\fasd,trim=0 0 0 0,clip]%
	    {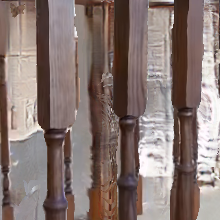}}\vspace{1mm}\\%
  \fbox{\includegraphics[width=\fasd,trim=0 0 0 0,clip]%
	    {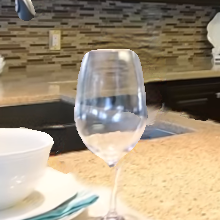}}\vspace{-1mm}\\%
  {Ours}
}\hfill%
\vspace{\figcapmargin}
\caption{\tb{Failure cases.} Single-image depth estimation algorithms (e.g., MegaDepth) often have difficulty in handling thin and complex structures and may produce overly smooth depth maps.
}

\label{fig:failure}
\end{figure*}

\subsection{Failure modes.} 
Our method uses an explicit depth map representation.
Consequently, the proposed approach does not handle reflective/transparent surfaces or complex/thin occluders very well.
\figref{failure} shows several failure cases.
}

\ignorethis{
\textcolor{blue}{Next, we adopt the testing protocal in ~\cite{zhou2018stereo} to compare with other works quantitatively. Also, an ablation study focuses on different kinds of inpainting procedure is conducted. Furthermore, we leverage AMT to evaluate the visual quality of 3D Photo virtual exploration. Finally, we show sample visual result compared with other works.}



\subsection{Implementation details} 
\heading{Image preprocessing.} 
\textcolor{blue}{We perform parameter sweeping on five withheld (validation) images and the examine the visual quality of novel views. We empirically set the threshold of disparity difference as 0.04 for depth maps from iPhone X and RealEstate10K (when the disparity map is normalized to [0,1])}

\heading{Training the inpainting model.} For the edge-generator, we follow the hyper-parameters in~\cite{nazeri2019edgeconnect}. Specifically, we train the edge-generator model using the ADAM optimizer~\cite{kingma2014adam} with $\beta=0.9$ and with an initial learning rate of $0.0001$. 
We train both the edge and depth generator model using the context-synthesis regions dataset on the MS-COCO dataset for 5 epochs. We train the depth generator and color image generator for 5 and 10 epochs, respectively. 

\heading{Inpainting model architecture.} For the edge inpainting network, we adopt the architecture provided by~\cite{nazeri2019edgeconnect}. For the depth and color inpainting networks, we use a standard U-Net architecture with partial covolution~\cite{liu2018image}. 

Due to the space limitation, we leave additional implementation details (specific network architecture, the training loss and their weights for each network) to the supplementary material. We will make the source code and pre-trained model publicly available to foster future work.

\begin{figure}[t]
\centering
\includegraphics[width=0.8\linewidth]{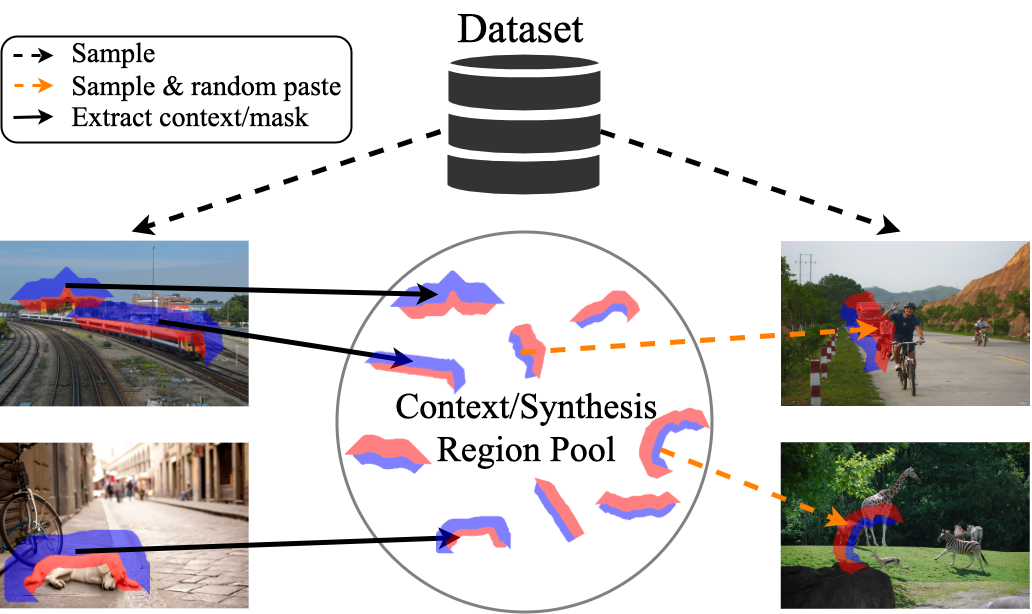}
\caption{
\tb{Dataset generation process.} We first form a collection of context/synthesis regions by extracting them from the linked depth edges in images on the COCO dataset. We then randomly sample and paste these regions onto \emph{different} images, forming our training dataset for context-aware color and depth inpainting.
}%
\label{fig:coco_dataset}
\end{figure}
\heading{Training data.}
While prior work requires specific data type, such as multi-view posed dataset~\cite{tulsiani2018layer} or video data~\cite{zhou2018stereo} for training, our proposed can be simply trained on any image dataset without the need of annotated data. Here, we choose to use MSCOCO dataset~\cite{lin2014microsoft} for its wide diversity in object types and scenes. To generate the training data for the inpainting model, we first use MegaDepth~\cite{li2018megadepth} to infer depth maps from MSCOCO images, and then process the estimated depth maps to generate context-synthesis regions. We augment our training data by randomly pasting the context-synthesis pairs to other images.  (See \figref{coco_dataset})


\begin{table}[t]
\caption{\tb{Quantitative comparison} on the RealEstate10K dataset.}
\label{tab:quantitative_other}
\centering
\small
\begin{tabular}{l|ccc}
\toprule
    Methods &   SSIM $\uparrow$    &   PSNR $\uparrow$   &   LPIPS $\downarrow$    \\ 
    \midrule
    Stereo-Mag~\cite{zhou2018stereo}   &  \tb{0.8906} & 	26.71 &	0.0826   \\ 
    PB-MPI~\cite{srinivasan2019pushing}   &     0.8773    &   25.51   & 0.0902   \\ 
    LLFF~\cite{mildenhall2019llff}   &   0.8697   &   24.15   &   0.0941   \\ 
    Xview~\cite{choi2019extreme}   &   0.8628   &   24.75   &   0.0822   \\ 
    Ours   &   0.8887       &   \tb{27.29}  &  \tb{0.0724}     \\ 
    \bottomrule
\end{tabular}
\end{table}

\subsection{Quantitative Evaluation}
\heading{Comparisons to recent works} We compare our method with recent works~\cite{zhou2018stereo} \cite{choi2018extreme} \cite{mildenhall2019llff} by 1500(?)[Currently 400] sampled triplets from RealEstate10K.
(See ~\tabref{quantitative_other})

\newcommand{\qn}[1]{(\blue{#1})}
\begin{table}[t]
\caption{\tb{ Using depth edge as guidance improves the results.} \blue{Blue}: results in disocculded regions.}
\label{tab:ablation}
\centering
\setlength{\tabcolsep}{3pt}
\small
\begin{tabular}{l|ccc}
\toprule
    Methods &   SSIM $\uparrow$    &   PSNR $\uparrow$   &   LPIPS $\downarrow$    \\ 
\midrule
    Diffusion   &   0.8665 \qn{0.6237}   &   25.95 \qn{18.91}   &   0.084   \\ 
    Inpaint w/o edge   &   0.8665 \qn{0.6247} &   25.96 \qn{18.94}   &   0.084   \\ 
    Inpaint w/ edge (Ours)   &   0.8666 \qn{0.6265}   &   25.97 \qn{18.98}   &   0.083   \\ 
    \bottomrule
\end{tabular}
\end{table}

\heading{Ablation Study} We compare our method with different inpainting procedure. (See ~\tabref{ablation}) The first setting is diffuse the depth from context region. We let the depth value in context region flux into the occlusion region. The second setting is to use a depth inpainting model to complete the occlusion area without the guidance of inpainted depth edge. The third setting is to inpaint depth map in occlusion area with the guidance of inpainted depth edge.

\begin{table}[t]
\caption{Human evaluation to recent works.}
\label{tab:human_other}
\vspace{\tabcapmargin}
\begin{center}
\begin{tabular}{c|c}
    Methods &   Our \\ \hline
    LLFF   &   XX \% \\ 
    Stereo-Mag   &   XX \% \\ 
    PB-MPI   &   XX \% \\ 
    Extreme View &   XX \% \\ 
\end{tabular}
\end{center}
\end{table}
\heading{Human Evaluation on virtual exploration.} Generate video by free-viewpoint rendering is an essential feature in View Synthesis. But the metrics in previous section focuses on the comparison in single image. Hence, we do human evluation to account for artifacts that appears in the video. We generate virtual exploration videos from XX samples in RealEstate10K and ask user to select the better one. (See ~\tabref{human_other})

\subsection{Qualitative Result} Here we show sample visual results compared with recent methods. Additional examples and videos of free-viewpoint rendering can be found in the supp. material.

\subsection{Apply to other depth source} Here we show sample visual results with different source of depth map. (IPhone, Kinect, MiDaS).

\heading{Methods compared.} We select three representative methods for novel view synthesis: Stereo Magnification~\cite{zhou2018stereo}, Layered Scene Inference (LSI)~\cite{tulsiani2018layer}, and Facebook 3D photo.

\heading{Comparison to Stereo Magnification~\cite{zhou2018stereo}.} Stereo magnification takes a stereo pair of images as input. 
Here, we use the examples provided by~\cite{zhou2018stereo} but use only the right color image. We obtain the corresponding depth maps of these images using MegaDepth~\cite{li2018megadepth}. We use the aligned color/depth images as inputs for our algorithm. To obtain the results of stereo magnification, we use the pre-trained model to produce multi-plane images for each testing ex\-am\-ple and use them to render the scene at novel viewpoints\footnote{Available at \url{https://github.com/google/stereo-magnification}}.

\figref{qualitative_mpi} shows two examples of challenging scenes.  While the stereo magnification method produces plausible view extrapolation results along the camera baseline direction, the synthesized novel views reveal clearly visible artifacts, particularly along the object boundaries. 



\heading{Comparison with Layered Scene Inference~\cite{tulsiani2018layer}.} As our method shares similar 3D representation as \cite{tulsiani2018layer} (layered depth images), we show visual comparison to highlight the importance of performing the proposed localize inpainting with spatially adaptive contexts. As the model from LSI~\cite{tulsiani2018layer} is \emph{domain-specific}, we evaluate the pre-trained model provided by the authors on images from the KITTI raw dataset. 

\newlength\qlsa
\setlength\qlsa{4.10cm}
\newlength\qlsb
\setlength\qlsb{1.0mm}

\def\qlaa{0.3153988868274583}
\def\qlab{0.2565217391304348}
\def\qlac{0.3196969696969697}

\begin{figure}%
\centering%
\parbox[t]{\qlsa}{\vspace{0mm}\centering%
  \includegraphics[width=\qlsa,height=\qlaa\qlsa]%
	    {figure/qualitative_lsi/LSI_ex1_baseline}\vspace{\qlsb}\\%
  \includegraphics[width=\qlsa,height=\qlab\qlsa]%
	    {figure/qualitative_lsi/LSI_ex2_baseline}\vspace{\qlsb}\\%
  \includegraphics[width=\qlsa,height=\qlac\qlsa]%
	    {figure/qualitative_lsi/LSI_ex4_baseline}\vspace{-1mm}\\%
	LSI results~\cite{tulsiani2018layer}}%
\hfill%
\parbox[t]{\qlsa}{\vspace{0mm}\centering%
  \includegraphics[width=\qlsa,height=\qlaa\qlsa]%
	    {figure/qualitative_lsi/LSI_ex1_ours}\vspace{\qlsb}\\%
  \includegraphics[width=\qlsa,height=\qlab\qlsa]%
	    {figure/qualitative_lsi/LSI_ex2_ours}\vspace{\qlsb}\\%
  \includegraphics[width=\qlsa,height=\qlac\qlsa]%
	    {figure/qualitative_lsi/LSI_ex4_ours}\vspace{-1mm}\\%
	Our results}\vspace{1mm}\\%
\caption{Qualitative results compare to LSI~\cite{tulsiani2018layer}.}
\label{fig:qualitative_lsi}

\end{figure}

\figref{qualitative_lsi} shows visual comparisons with two examples. While our model is \emph{domain-agnostic} (e.g., our inpainting model has \emph{not} trained on street view images), our results contain significantly fewer artifacts.

\heading{Comparison with Facebook 3D photo.} Here, we aim to evaluate the capability of our method on photos taken \emph{in the wild}. We extract the color images and the corresponding depth maps estimated from an iPhone X (with dual camera lens). We use the same set of RGB-D inputs for both Facebook 3D photo and our algorithm. 

 shows the view synthesis results in comparison with Facebook 3D photo. The diffused color and depth values by the facebook 3D photo algorithm work well when small or thin occluded regions are revealed at the novel view. These artifacts, however, become clearly visible with larger occluded regions. On the other hand, our results in general fills in the synthesis regions with visually plausible contents and structures.


\subsection{Ablation study}
We investigate the importance of our model design and their impacts on the visual quality of our results. Specifically, building upon our full model, we test two variants: 
1) replacing the color inpainting with diffused colors and 
2) inpainting the color and depth without inferring the depth discontinuities in the occluded regions. 

\figref{qualitative_ablation} shows one example of these variants. 
Without color inpainting, the model cannot produce realistic novel views when the background is complex. 
Without the estimated depth discontinuities, our results show artifacts of stretched meshes along the object boundaries. 

}



\ignorethis{
\subsection{Failure modes}

While we have shown substantial improvement over the state-of-the-art, free-viewpoint novel view synthesis remains a challenging problem. Here we discuss several failure modes of our algorithm. 
Example 1: Inaccurate depth

Example 2: Complex depth structures (tree)

Example 3: Inpainted color are bad
}
\vspace{-2mm}

\section{Conclusions}
\label{sec:conclusions}

In this paper, we present an algorithm for creating compelling 3D photography from a single RGB-D image. 
Our core technical novelty lies in creating a completed layered depth image representation through context-aware color and depth inpainting. 
We validate our method on a wide variety of everyday scenes. 
Our experimental results show that our algorithm produces considerably fewer visual artifacts when compared with the state-of-the-art novel view synthesis techniques. 
We believe that such technology can bring 3D photography to a broader community, allowing people to easily capture scenes for immersive viewing.



\vspace{-1mm}
{\flushleft {\bf Acknowledgement.}} This project is supported in part by NSF (\#1755785) and MOST-108-2634-F-007-006 and MOST-109-2634-F-007-016.

{\small
\bibliographystyle{ieee_fullname}
\bibliography{main}
}

\onecolumn
{
\Large
\begin{center}
\textbf{\noindent 3D Photography using Context-aware Layered Depth Inpainting\\Supplementary Material}
\end{center}
}





\begin{table}[h]
\caption{\tb{Quantitative comparison} on the RealEstate10K dataset.}
\label{tab:quantitative_errata}
\centering
\small
\begin{tabular}{l|ccc}
\toprule
    Methods &   SSIM $\uparrow$    &   PSNR $\uparrow$   &   LPIPS $\downarrow$    \\ 
    \midrule
    Stereo-Mag~\cite{zhou2018stereo}   &  \tb{0.8906} & 	26.71 &	0.0826   \\ 
    PB-MPI~\cite{srinivasan2019pushing} (32 Layers)   &     0.8717    &   25.38   & 0.0925   \\ 
    PB-MPI~\cite{srinivasan2019pushing} (64 Layers)   &     0.8773    &   25.51   & 0.0902   \\ 
    PB-MPI~\cite{srinivasan2019pushing} (128 Layers)   &     0.8700    &   24.95   & 0.1030   \\ 
    LLFF~\cite{mildenhall2019llff}   &   0.8697   &   24.15   &   0.0941   \\ 
    Xview~\cite{choi2019extreme}   &   0.8628   &   24.75   &   0.0822   \\ 
    Ours   &   0.8887       &   \tb{27.29}  &  \tb{0.0724}     \\ 
    \bottomrule
\end{tabular}
\end{table}

\newcommand{\norm}[1]{\vert\vert#1\vert\vert}
\section{Additional Quantitative Results}
We further evaluate the PB-MPI method~\cite{srinivasan2019pushing} with various number of depth layers. We report the results in~\tabref{quantitative_errata}. 

\ignorethis{\paragraph{Quantitative comparison (Section 4.3)} After the submission, we discover that a small fraction of testing triplets are corrupted. We manually go through all the 1386 samples in our testing set, remove 88 corrupted triplets, and re-compute the performance metrics with the remaining 1298 testing triplets. 
\tabref{quantitative_errata} shows the result on the \emph{cleaned} testing set. 

Overall, we arrive at the same conclusion as in the main paper: our method performs competitively on PSNR and SSIM, and exhibit better perceptual quality as reflected in the LPIPS score.
In addition, we report two additional novel view synthesis methods on this testing set: Local Light Field Fusion~\cite{mildenhall2019llff} and Extreme View Synthesis~\cite{choi2019extreme} using the code provided by the authors.}

\section{Visual Results}

\heading{Comparisons with the state-of-the-arts.} 
We provide a collection of rendered 3D photos with comparisons with the state-of-the-art novel view synthesis algorithms. 
In addition, we show that our method can synthesize novel view for legacy photos. 
Please refer to the website\footnote{\label{note1}\url{https://shihmengli.github.io/3D-Photo-Inpainting/}} for viewing the results.

\heading{Ablation studies.}
To showcase how each of our proposed component contribute to the quality of the synthesized view, we include a set of rendered 3D photos using the same ablation settings in Section 4.4 of the main paper. 
Please refer to the website\textsuperscript{\ref{note1}} for viewing the photos.

\section{Implementation Details}
In this section, we provide additional implementation details of our model, including model architectures, training objectives, and training dataset collection.
We will release the source code to facilitate future research in this area.

\begin{table}[t]
\caption{\tb{Model architecture of our color and depth inpainting models.} W denote partial convolution layer as PConv, and denote BatchNorm as BN. We add the context and synthesis region together as the partial masks for the PConv layers.}%
\label{tab:arch_color_depth}
\centering
\small
\begin{tabular}{ccccccc}
\toprule
Module&   Filter Size &  \#Channels & Dilation  & Stride & Norm & Nonlinearity  \\ 
\midrule
PConv1 & $7\times7$ & 64 & 1 & 2 & - & ReLU\\
PConv2 & $5\times5$ & 128 & 1 & 2 & BN & ReLU\\
PConv3 & $5\times5$ & 256 & 1 & 2 & BN & ReLU\\
PConv4 & $3\times3$ & 512 & 1 & 2 & BN & ReLU\\
PConv5 & $3\times3$ & 512 & 1 & 2 & BN & ReLU\\
PConv6 & $3\times3$ & 512 & 1 & 2 & BN & ReLU\\
PConv7 & $3\times3$ & 512 & 1 & 2 & BN & ReLU\\
PConv8 & $3\times3$ & 512 & 1 & 2 & BN & ReLU\\
\midrule
NearestUpsample & - & 512 & - & 2 & - & - \\
Concatenate (w/ PConv7) & - & 512+512 & - & - & - & - \\
PConv9 & $3\times3$ & 512 & 1 & 1 & BN & LeakyReLU(0.2)\\
\midrule
NearestUpsample & - 512 & - & 2 & - & - \\
Concatenate (w/ PConv6) & - & 512+512 & - & - & - & - \\
PConv10 & $3\times3$ & 512 & 1 & 1 & BN & LeakyReLU(0.2)\\
\midrule
NearestUpsample & - & 512 & - & 2 & - & - \\
Concatenate (w/ PConv5) & - & 512+512 & 1 & - & - & - \\
PConv11 & $3\times3$ & 512 & 1 & 1 & BN & LeakyReLU(0.2)\\
\midrule
NearestUpsample & - & 512 & - & 2 & - & - \\
Concatenate (w/ PConv4) & - & 512+512 & - & - & - & - \\
PConv12 & $3\times3$ & 512 & 1 & 1 & BN & LeakyReLU(0.2)\\
\midrule
NearestUpsample & - & 512 & - & 2 & - & - \\
Concatenate (w/ PConv3) & - & 512+256 & - & - & - & - \\
PConv13 & $3\times3$ & 256 & 1 & 1 & BN & LeakyReLU(0.2)\\
\midrule
NearestUpsample & - & 256 & - & 2 & - & - \\
Concatenate (w/ PConv2) & - & 256+128 & - & - & - & - \\
PConv14 & $3\times3$ & 128 & 1 & 1 & BN & LeakyReLU(0.2)\\
\midrule
NearestUpsample & - & 128 & - & 2 & - & - \\
Concatenate (w/ PConv1) & - & 128+64 & - & - & - & - \\
PConv15 & $3\times3$ & 64 & 1 & 1 & BN & LeakyReLU(0.2)\\
\midrule
NearestUpsample & - & 64 & - & 2 & - & - \\
Concatenate (w/ Input) & - & 64 + 4 or 64 + 6 (Depth / Color Inpainting) & - & - & - & - \\
PConv16 & $3\times3$ & 1 or 3 (Depth / Color Inpainting) & 1 & 1 & - & - \\
\bottomrule
\end{tabular}
\end{table}
\begin{table}[t]
\caption{\tb{Model architecture of our edge inpainting models.} As in~\cite{nazeri2019edgeconnect}, the edge inpainting model consists of 1 edge generator, and 1 discriminator network. SN$\rightarrow$IN indicates that we first perform spectral normalization (SN)~\cite{miyato2018spectral}, and then apply instance normalization (IN)~\cite{ulyanov2016instance}. ResnetBlock comprises 2 conv layers with the specified hyper-parameters and a skip connection between the input and the output of the block.}%
\label{tab:arch_edge}
\centering
\small
\begin{tabular}{ccccccc}
\toprule
\multicolumn{7}{c}{\textbf{Edge Generator}} \\
Module&   Filter Size & \#Channels & Dilation & Stride & Norm & Nonlinearity  \\ 
\midrule
Conv1 & $7\times7$ & 64 & 1 & 1 & SN$\rightarrow$IN & ReLU \\
Conv2 & $4\times4$ & 128 & 1& 2 & SN$\rightarrow$IN & ReLU \\
Conv3 & $4\times4$ & 256 & 1 & 2 & SN$\rightarrow$IN & ReLU \\
\midrule
ResnetBlock4 & $3\times3$ & 256 & 2 & 1 & SN$\rightarrow$IN & ReLU \\
ResnetBlock5 & $3\times3$ & 256 & 2 & 1 & SN$\rightarrow$IN & ReLU \\
ResnetBlock6 & $3\times3$ & 256 & 2 & 1 & SN$\rightarrow$IN & ReLU \\
ResnetBlock7 & $3\times3$ & 256 & 2 & 1 & SN$\rightarrow$IN & ReLU \\
ResnetBlock8 & $3\times3$ & 256 & 2 & 1 & SN$\rightarrow$IN & ReLU \\
ResnetBlock9 & $3\times3$ & 256 & 2 & 1 & SN$\rightarrow$IN & ReLU \\
ResnetBlock10 & $3\times3$ & 256 & 2 & 1 & SN$\rightarrow$IN & ReLU \\
ResnetBlock11 & $3\times3$ & 256 & 2 & 1 & SN$\rightarrow$IN & ReLU \\
\midrule
ConvTranspose12 & $4\times4$ & 128 & 1 & 2 & SN$\rightarrow$IN & ReLU \\
ConvTranspose13 & $4\times4$ & 64 & 1 & 2 & SN$\rightarrow$IN & ReLU \\
Conv14 & $7\times7$ & 1 & 1 & 1 & SN$\rightarrow$IN & Sigmoid \\
\midrule
\toprule
\multicolumn{7}{c}{\textbf{Discriminator}} \\
Module&   Filter Size & \#Channels & Dilation & Stride & Norm & Nonlinearity  \\ 
\midrule
Conv1 & $4\times4$ & 64 & 1 & 2 & SN & LeakyReLU(0.2) \\
Conv2 & $4\times4$ & 128 & 1 & 2 & SN & LeakyReLU(0.2) \\
Conv3 & $4\times4$ & 256 & 1 & 2 & SN & LeakyReLU(0.2) \\
Conv4 & $4\times4$ & 512 & 1 & 1 & SN & LeakyReLU(0.2) \\
Conv5 & $4\times4$ & 1 & 1 & 1 & SN & Sigmoid \\
\bottomrule
\end{tabular}
\end{table}
\begin{table}[t]
\caption{\tb{Input of each model in our proposed method.} The check mark $\checkmark$ indicates that it is used as input for the model.}%
\label{tab:input_model}
\centering
\small
\begin{tabular}{lcccc}
\toprule
     &   RGB  &   Depth   &   Edge & Context\& Synthesis   \\ 
\midrule
    Color Inpainting   &   \checkmark  &   -   &   \checkmark & \checkmark   \\ 
    Depth Inpainting   &   -   &   \checkmark   &   \checkmark & \checkmark   \\ 
    Edge Inpainting   &   \checkmark   &  \checkmark   &  \checkmark & \checkmark   \\ 
    \bottomrule
\end{tabular}
\vspace{-4mm}
\end{table}

\paragraph{Model architectures.} We adopt the same U-Net~\cite{ronneberger2015u} architecture as in~\cite{liu2018image} for our depth inpainting and color inpainting models (see \tabref{arch_color_depth}), and change the input channels for each model accordingly. 
For the edge inpainting model, we use a design similar to~\cite{nazeri2019edgeconnect} (see \tabref{arch_edge}). 
We set the input depth and RGB values in the synthesis region to zeros for all three models. 
The input edge values in the synthesis region are similarly set to zeros for depth and color inpainting models, but remain intact for the edge inpainting network. 
We show the input details of each model in~\tabref{input_model}

\begin{figure}[t]
\centering
\includegraphics[width=0.8\linewidth]{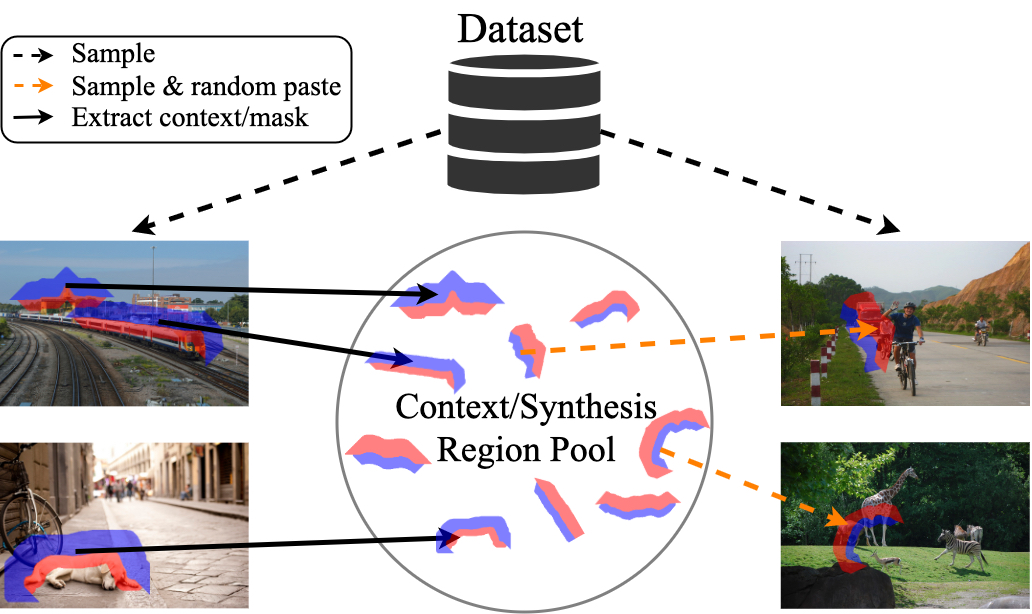}
\caption{
\tb{Dataset generation process.} We first form a collection of context/synthesis regions by extracting them from the linked depth edges in images on the COCO dataset. We then randomly sample and paste these regions onto \emph{different} images, forming our training dataset for context-aware color and depth inpainting.
}%
\label{fig:coco_dataset}
\end{figure}
\paragraph{Training objective.} To train our color inpainting model, we adopt similar objective functions as in ~\cite{liu2018image}. 
First, we define the reconstruction loss for context and synthesis regions:
\begin{align}
    L_\mathrm{synthesis}=\frac{1}{N}\norm{S\odot(I-I_{gt})},~~~~~~L_\mathrm{context}=\frac{1}{N}\norm{C\odot(I-I_{gt})},
\end{align}
where $S$ and $C$ are the binary mask indicating \emph{synthesis} and \emph{context} regions, respectively, $\odot$ denotes the Hadamard product, $N$ is the total number of pixels, $I$ is the inpainted result, and $I_{gt}$ is the ground truth image.

Next, we define the perceptual loss~\cite{johnson2016perceptual}:
\begin{equation}
    L_{perceptual}=\sum^{P-1}_{p}\frac{\norm{\psi_p(I)-\psi_p(I_{gt})}}{N_{\psi_p}},
\end{equation}
Here, $\psi_p(\cdot)$ is the output of the $p$th layer from VGG-16~\cite{simonyan2014very}, and $N_{\psi_p}$ is the total number of elements in $\psi_p(\cdot)$.

We define the style loss as:
\begin{equation}
    L_{style}=\sum^{P-1}_{p}\frac{1}{C_pC_p}\norm{\frac{1}{C_pH_pW_p}\left[(\psi^I_p)^{\top}\psi^I_p-(\psi^{I_{gt}}_p)^{\top}\psi^{I_{gt}}_p\right]},
\end{equation}
where $C_p$, $H_p$, $W_p$ is the number of channels, height, and width of the output $\psi_p(\cdot)$. 

Finally, we adopt the Total Variation (TV) loss:
\begin{equation}
    L_{tv}=\sum_{(i,j)\in S,(i,j+1)\in S}\frac{\norm{I(i, j+1)-I(i,j)}}{N}+\sum_{(i,j)\in S,(i+1,j)\in S}\frac{\norm{I(i+1, j)-I(i,j)}}{N}.
\end{equation}

Here, We overload the notation $S$ to denote the synthesis region. 
This term can be interpreted as a smoothing penalty on the synthesis area.
Combine all these loss terms, we obtain the training objective for our color inpainting model:
\begin{equation*}
    L = L_{context} + 6L_{synthesis}+0.05L_{perceptual}+120L_{style}+0.01L_{tv}
\end{equation*}
For our depth inpainting model, we use only $L_{context} + L_{synthesis}$ as the objective functions. 
For edge inpainting model, we follow the identical training protocol as in~\cite{nazeri2019edgeconnect}.

\paragraph{Training details.} 
We illustrate the data generation process in~\figref{coco_dataset}. 
We use the depth map predicted by MegaDepth~\cite{li2018megadepth} as our pseudo ground truth.
We train our method using 1 Nvidia V100 GPU with batch size of 8, and the total training time take about 5 days.

\ignorethis{
\section{User Study}
We conduct subjective test to evaluate the performance of our proposed method. To do so, we generate a total \red{N} pairs of 3D photos. Each pair consists one 3D photo generated from our method, and one generated by either Stereo Magnification~\cite{zhou2018stereo} or Facebook 3D photo. The participants of our subjective test were asked to select the one with better visual quality from each pair. We show the evaluation results in Table~\ref{}. Overall, the results indicate that our method is preferred by the majority of participants to prior methods.
}

\ignorethis{
\input{figure/fig_.tex}
\section{ Study}
We investigate how the color inpainting and depth edge inpainting model affect our performance of our method. \figref{qualitative_} shows the outcomes. 
Our result show that the depth edge inpainting model plays an critical role for filling in incomplete structures in the occluded regions.
Removing the depth edge inpainting model results in several visible holes (shown as white regions) when rendered at novel views.
The use of context-aware color inpainting model allows us to synthesize plausible contents. We show that replacing the color inpainting model with simple diffusion methods produces blurry results.
Both depth edge and color inpainting models help improve the visual quality of our results.
}

\section{Failure cases}
As estimating depth/disparity map from a single image remain a challenging problem (particularly for scenes with complex, thin structures), our method fails to produce satisfactory results with plausible motion parallax for scenes with complex structures. 
Due to the use of explicit depth map, our method is unable to handle reflective/transparent surfaces well. 
We show in \figref{failure} two examples of such cases. Here, we show the input RGB image as well as the estimated depth map from the pre-trained MegaDepth model. 
The rendered 3D photos can be found in the supplementary webpage.

%

\clearpage

\end{document}